\documentclass[]{amap}
\usepackage{amssymb}

\usepackage[toc,page,header]{appendix}
\usepackage{minitoc}
\usepackage{solarized-light}
\usepackage{booktabs}
\usepackage{multirow}
\usepackage{graphicx}
\usepackage{array}
\usepackage{siunitx,array}
\usepackage[table]{xcolor}
\usepackage{makecell,array,booktabs}
\usepackage{makecell}
\usepackage{framed}

\usepackage{bbding}
\usepackage{hyperref}
\usepackage{amsmath}
\usepackage{amsfonts}
\usepackage{placeins}
\usepackage[colorinlistoftodos]{todonotes}
\usepackage{longtable}
\usepackage{hhline}
\usepackage{fancyvrb}
\usepackage{graphicx}
\usepackage[table]{xcolor}
\usepackage{booktabs}
\usepackage{float}
\usepackage{fvextra}
\usepackage{CJKutf8}
\usepackage{multicol}
\usepackage{float}
\usepackage{placeins}
\usepackage{cleveref}
\usepackage{tablefootnote}
\usepackage{threeparttable}
\usepackage{tabularx}
\usepackage{mdframed}
\usepackage{subcaption}
\usepackage[usestackEOL]{stackengine}
\usepackage[numbers]{natbib}
\newcommand{\commentout}[1]{}
\renewcommand{\paragraph}[1]{\noindent\textbf{#1.}\hspace*{1em}}
\usepackage{enumitem}
\usepackage{hyperref}
\setlist[itemize]{leftmargin=15pt}
\usepackage{siunitx,array}
\usepackage{tabularx}
\usepackage{adjustbox}
\usepackage{arydshln}
\usepackage{pifont}
\usepackage{bbding}
\definecolor{ampblue}{rgb}{0.82, 0.88, 0.94}
\usepackage[table]{xcolor}
\usepackage{array}
\usepackage[dvipsnames]{xcolor}

\RequirePackage{xspace}
\makeatletter
\DeclareRobustCommand\onedot{\futurelet\@let@token\@onedot}
\def\@onedot{\ifx\@let@token.\else.\null\fi\xspace}

\def\eg{\emph{e.g}\onedot} 
\def\ie{\emph{i.e}\onedot}

\makeatother

\newcommand{\method}{\textit{ABot-3DWorld 0}\xspace}

\newcommand{\abotgs}{ABot-3DGS\xspace}
\newcommand{\spatialprim}{SGP\xspace}
\newcommand{\Plucker}{Pl{\"u}cker}

\crefname{figure}{Fig.}{Figs.}
\Crefname{figure}{Fig.}{Figs.}
\crefname{table}{Tab.}{Tabs.}
\Crefname{table}{Tab.}{Tabs.}
\Crefname{section}{Sec.}{Secs.}
\Crefname{section}{Sec.}{Secs.}

\usepackage{xcolor}

\definecolor{abot1}{HTML}{0185FE}
\definecolor{abot2}{HTML}{0185FE}
\definecolor{abot3}{HTML}{0185FE}
\definecolor{abot4}{HTML}{0185FE}
\definecolor{abot5}{HTML}{FB8C00}
\definecolor{abot6}{HTML}{FB8C00}
\definecolor{abot7}{HTML}{FB8C00}

\title{ABot-3DWorld 0: A Universal World Model to\\Explore Any 3D Space}

\author{AMAP CV Lab}
\vspace{-40pt}

\abstract{
We present \method{}, a universal multimodal 3D world model that turns text, image, and video inputs into high-fidelity, explorable 3D worlds. At the heart of our framework is a unified Spatial Generative Primitive (SGP)---a compact tuple of a high-quality panorama and a spatial point cloud that delivers an efficient description of any 3D space. Multimodal inputs are first lifted into this primitive; a 3D-consistent panoramic video generator then explores the primitive along a planned trajectory; finally, our panoramic video reconstruction engine converts the generated video into a clean, photorealistic 3D Gaussian Splatting (3DGS) world. This pipeline covers two regimes: rich inputs (multi-view sets, casual video) are lifted into the SGP through a geometry-rigorous recovery that mirrors the observed scene, while a single image or sentence is completed generatively into a creative world. The result is one low-barrier engine for general 3D content creation that further anchors generated worlds to geographic points of interest, enabling map-native spatial exploration at consumer scale. Experiments show that \method{} sets the state of the art among open-source methods and demonstrates stronger scene fidelity than Marble under rich multimodal inputs.

\bigskip

\textbf{Official Page: } \url{https://abot-world.amap.com/plaza}
}

\begin{document}
\maketitle
\vspace{-4pt}

\begin{figure}[H]
\centering
\includegraphics[width=\textwidth]{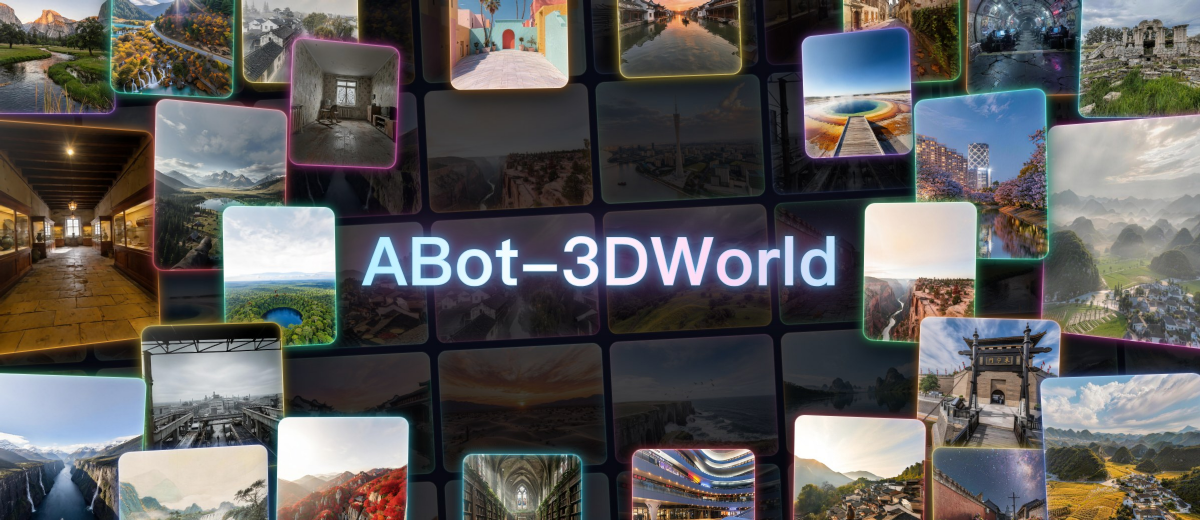}
\vspace{-6pt}
\caption{Explorable 3DGS worlds generated by \method{} from diverse inputs (text, a single image, multiview photos, or video), spanning indoor, street-level, and aerial scenes, from faithful reconstructions to fantastical creations---all produced by a single unified pipeline.}
\label{fig:teaser}
\end{figure}

\clearpage

\section{Introduction}
\label{sec:intro}

AI-driven 3D content creation is undergoing a progression from objects to worlds. Object-level 3D generation, \ie{} synthesizing a single 3D asset from text or a single image via score distillation~\cite{poole2022dreamfusion} or native mesh generation, has already spawned multiple commercialized companies and products (\eg{} Tripo, Meshy, Tencent Hunyuan3D~\cite{hunyuan3d2025}), validating the efficiency and quality of direct 3D synthesis. Its application scope, however, stops at individual assets. Scene generation directly enables immersive spatial experiences, virtual tourism, and scalable game-world production, and offers substantially greater application value through product forms that object generation alone cannot support. Scaling from an isolated object to a complete, explorable 3D scene dramatically increases spatial complexity and scale, giving rise to two core technical challenges. First, how to efficiently represent a complete space: an object can be fully covered by a single circular sweep, whereas a scene extends from a single room to a streetscape or an entire city block, demanding a compact representation that simultaneously encodes geometric structure and visual appearance. Second, how to efficiently explore that space: the narrow field of view of perspective imagery means that covering a full environment requires a large number of frames, with generation and rendering costs growing sharply as scene scale increases. Additionally, scene-specific phenomena such as sky, water surfaces, and long-range illumination introduce difficulties absent from object capture. These challenges explain why only a handful of systems~\cite{yu2024luciddreamer,fridman2024scenescape,yu2024wonderworld} have attempted full 3D scene generation to date.

Recent systems such as Marble~\cite{marble2025worldlabs} and HY-World 2.0~\cite{hyworld22026} have demonstrated that generative models can already transform flat inputs into navigable 3D scenes expressed as neural radiance fields~\cite{mildenhall2020nerf}, 3D Gaussian Splatting~\cite{3DGS}, or polygonal meshes, substantially advancing the state of the art. Yet several fundamental technical bottlenecks remain open: (i) multimodal inputs (multi-view images, video) are often processed through cascaded generative conversions that discard the geometric evidence present in the original observations and amplify hallucination, causing the generated scene to diverge from the input; (ii) video generators typically adopt perspective video as the spatial exploration carrier, but its narrow field of view leads to low coverage efficiency; moreover, these generators are optimized with 2D losses that lack cross-view geometric awareness, producing visually attractive but 3D-inconsistent outputs; and (iii) 3D scene generation places stringent demands on the geometric quality of training data---if training supervision comes solely from 2D data without aligned 3D ground truth, the 3D fidelity ceiling of the system is directly limited.

\begin{figure*}[!t]
\centering
\includegraphics[width=\textwidth]{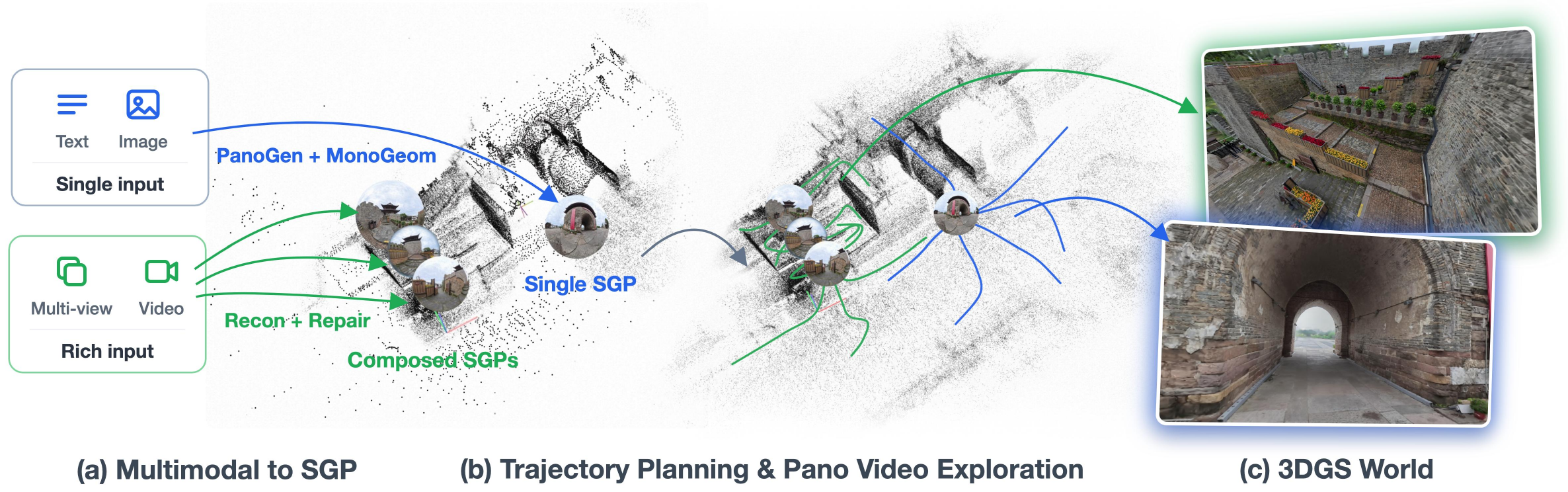}
\caption{Overview of the \method{} pipeline. \textbf{(a) Multimodal-to-SGP.} Every input modality is lifted into a Spatial Generative Primitive (SGP); rich inputs may compose into multiple SGPs. \textbf{(b) Trajectory planning and panoramic exploration.} A geometry-grounded planner and a 3DRL post-trained panoramic video generator explore the SGP. \textbf{(c) 3DGS world.} \abotgs{} with iterative FLUX repair reconstructs the final scene.}
\label{fig:pipeline_overview}
\end{figure*}

In this report we present \method{}, a universal multimodal 3D world model designed to address the above challenges within a single, unified pipeline (\cref{fig:pipeline_overview}). Our design centers on a four-stage architecture:

\begin{enumerate}[nosep,leftmargin=20pt]
\item \textbf{Multimodal input $\rightarrow$ Spatial Generative Primitive (SGP).} We define a compact canonical representation of 3D space: the SGP $\mathcal{S} = (\mathbf{P}, \mathbf{G})$, consisting of a high-quality panorama and a spatial point cloud. Every supported input modality (text, single image, multi-view photos, video) is first lifted into one or more SGPs through modality-appropriate pathways (\cref{sec:method_input}). Rich inputs are processed through a rigorous geometry pipeline to produce SGPs; minimal inputs are completed generatively via a DiT-based panorama generator (\cref{sec:method_pano}) followed by monocular geometry estimation~\cite{moge2}.

\item \textbf{SGP $\rightarrow$ exploration trajectory.} Given the SGP, a geometry-grounded agentic planner infers navigability from the point cloud, identifies semantically salient destinations via a vision-language model, and produces camera paths that jointly maximize spatial coverage, exploration efficiency, and reconstruction quality (\cref{sec:method_traj}).

\item \textbf{Trajectory $\rightarrow$ 3D-consistent panoramic video.} A 14B-parameter panoramic video generator built on VACE~\cite{vace2025} synthesizes an equirectangular exploration video conditioned on the SGP's point cloud. Three design choices are critical: dense point-cloud conditioning that grounds every frame in scene geometry; latent circular padding that eliminates the panoramic wrap-around seam; and 3D Reinforcement Learning (3DRL), a post-training stage that improves video quality and cross-view 3D consistency (\cref{sec:method_video,sec:method_3drl}).

\item \textbf{Panoramic video $\rightarrow$ 3DGS world.} The generated video is lifted into a clean, photorealistic 3DGS world by our \abotgs{} reconstruction engine. An iterative FLUX-based repair module super-resolves the inherently low-resolution panoramic frames and fixes residual view-inconsistent artifacts, while wild-scene robustness modules (sky separation, appearance calibration, MCMC densification, transient suppression) ensure stable reconstruction across all scene regimes (\cref{sec:method_gs}).
\end{enumerate}

Our main contributions are as follows:

\begin{itemize}
    \item \textbf{One pipeline for all modalities via the SGP.} The SGP unifies all input modalities into a single canonical representation, eliminating the need for scene-specific routing or specialized branches. Its rich-input path is what separates \method{} from competing systems: where others collapse multi-image inputs through generative image-to-panorama models and video inputs through video-to-text-to-panorama cascades, both discarding the original observations and amplifying hallucination, our geometry-rigorous route faithfully reconstructs the observed scene with minimal hallucination.

    \item \textbf{Panoramic video for efficient spatial exploration.} We make panoramic video, rather than perspective video or keyframe fusion, the canonical exploration medium of \method{}. Because each 360\textdegree{} frame observes the full surrounding environment, far fewer frames are needed to cover a given space compared with narrow field-of-view alternatives, making exploration inherently more efficient. The choice additionally delivers native spatio-temporal consistency and eliminates viewpoint blind spots. Its two known costs (no native 3D awareness, limited single-frame resolution) are closed by two compatible additions: (a) \textbf{3D Reinforcement Learning (3DRL)}, a post-training stage that improves video quality and cross-view 3D consistency; and (b) an \textbf{iterative FLUX-based 3DGS repair} module that super-resolves panoramic frames and removes residual view-inconsistent artifacts.

    \item \textbf{A full-coverage 3D asset pipeline for perfectly-aligned end-to-end training data.} The entire pipeline is powered by a purpose-built data engine (\cref{sec:data}) grounded in \abotgs{}, our full-spectrum 3DGS reconstruction system. \abotgs{} produces stable 3DGS scenes spanning indoor, street-level, and aerial environments from both proprietary captures and open-source datasets. A virtual panoramic camera then renders time-aligned (RGB, point-map, trajectory) tuples from these scenes, producing high-quality and diverse training data whose distribution is consistent between training and inference. This form of supervision is impossible to obtain from real panoramic captures, which inevitably suffer from camera shake, photographer occlusion inherent to 360\textdegree{} panoramic acquisition, and a lack of aligned geometric ground truth.
\end{itemize}

The remainder of this report is organized as follows. \cref{sec:data} presents the data engine that supplies aligned training supervision across all learnable modules. \cref{sec:method} details the core algorithmic components: SGP definition and multimodal extraction (\cref{sec:method_input}), panorama generation (\cref{sec:method_pano}), geometry-grounded trajectory planning (\cref{sec:method_traj}), panoramic video generation with 3DRL post-training (\cref{sec:method_video,sec:method_3drl}), \abotgs{}-based reconstruction and repair (\cref{sec:method_gs}), and physics-aware rendering extensions (\cref{sec:method_physics}). \cref{sec:eval} reports end-to-end comparisons against Marble and HY-World 2.0, and \cref{sec:conclusion} discusses limitations and future directions.

\section{Data}
\label{sec:data}

\begin{figure}[t]
\centering
\includegraphics[width=1.0\textwidth]{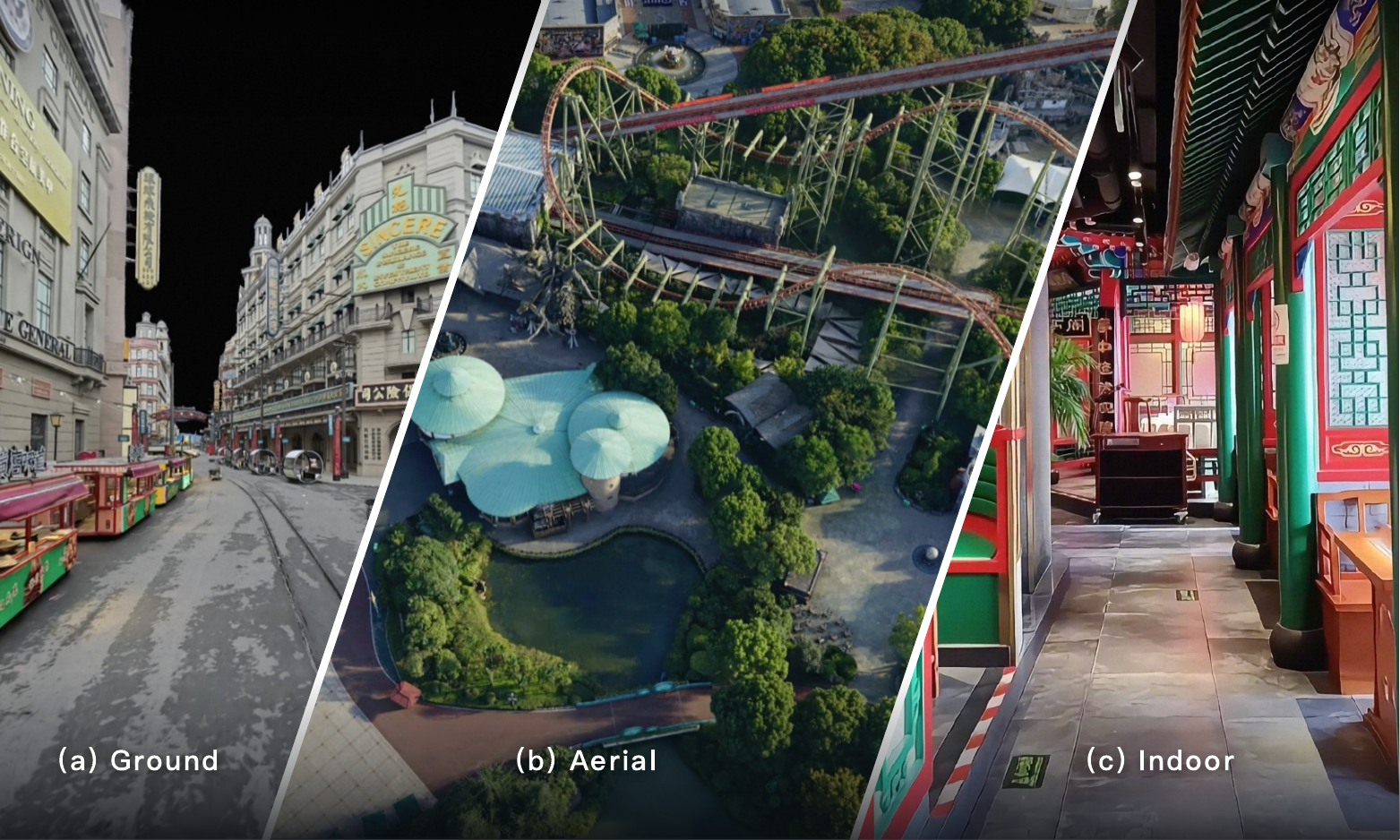}
\caption{Representative panoramic samples from our proprietary dataset, reconstructed by \abotgs{} across three scene categories: (a)~ground-level urban streets, (b)~aerial city views, and (c)~indoor architectural interiors.}
\label{fig:data_hero}
\end{figure}

\begin{figure}[t]
\centering
\includegraphics[width=\textwidth]{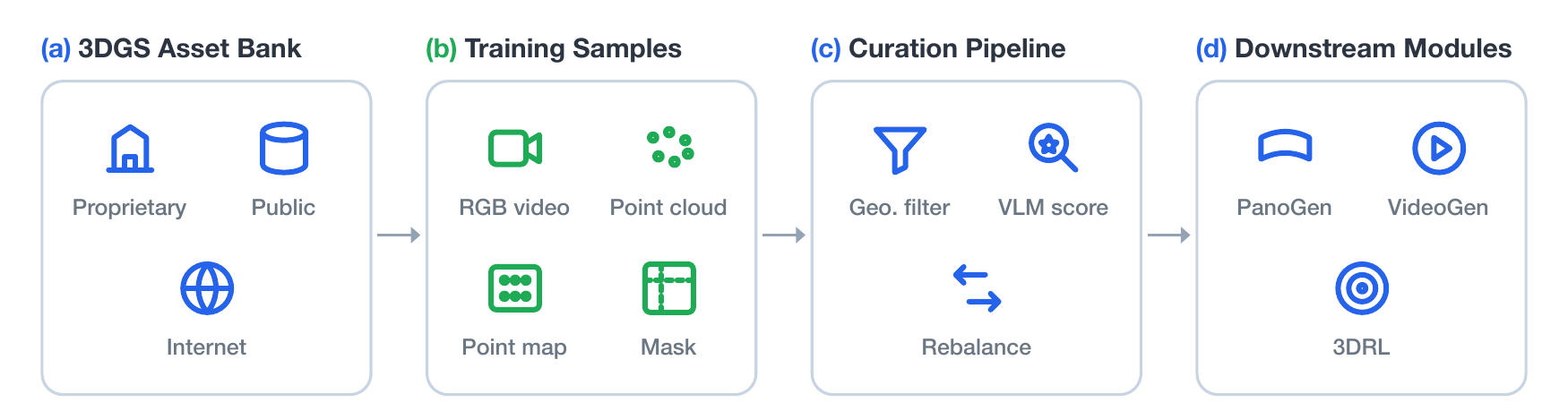}
\caption{The \method{} data engine. \textbf{(a)}~Multi-source assets are reconstructed into a unified 3DGS asset bank by \abotgs{}. \textbf{(b)}~A virtual panoramic camera renders aligned training samples (RGB video, point-cloud renderings, point maps, masks) along planned trajectories. \textbf{(c)}~A curation pipeline filters, scores, and rebalances the clips. \textbf{(d)}~The curated corpus supervises all downstream modules.}
\label{fig:data_pipeline}
\end{figure}

Several modules of \method{} require training on panoramic imagery or video: the panorama generator (\cref{sec:method_pano}), the panoramic video diffusion model (\cref{sec:method_video}), and the 3DRL post-training stage (\cref{sec:method_3drl}). Sourcing this supervision from real panoramic captures or public panorama collections is fundamentally limited by three properties of physical acquisition: (i) the operator, rig, or vehicle is unavoidably visible in the scene, corrupting the very geometry the model is meant to learn; (ii) the capture trajectory is not fully controllable and always carries residual camera shake, making it impossible to collect stable, controllable samples at arbitrary viewpoints or along specified trajectories; and (iii) no 3D geometric ground truth (depth, point map) is available, forcing supervision to fall back on noisy monocular pseudo-labels. We sidestep all three by adopting a capture-reconstruct-render workflow rather than using captured footage directly, so that every supervision signal is generated by rendering. Our data engine is grounded in \abotgs{}, a full-spectrum 3DGS reconstruction system that produces stable scenes spanning indoor rooms, street-level exteriors, and aerial cityscapes (\cref{fig:data_hero}). For each scene, a virtual panoramic camera then renders diverse trajectories, producing at every frame a (RGB, point-map, camera-pose) tuple that is, by construction, occlusion-free, fully pose-controllable, and perfectly geometry-aligned. Combined with the full data pipeline (\cref{fig:data_pipeline}) that further performs deduplication, cleansing, and category rebalancing, this yields a high-quality 3D panoramic dataset that supports the training of the modules listed above.

\subsection{Building the 3DGS Asset Bank}
\label{sec:data_assets}

The first component of our data engine is a curated bank of high-quality 3DGS scenes that we treat as a continuously growing ``virtual world''. We build this bank by fusing proprietary AMAP assets with open-source datasets and internet-sourced material, drawing from four complementary sources: proprietary assets carry the three core scene categories (indoor rooms, street-level exteriors, and aerial cityscapes), while open-source datasets and internet material respectively supply real-world indoor variation and a stylistic long tail.

\paragraph{Proprietary indoor and outdoor 3DGS scenes}
We use internally captured indoor and outdoor video sequences, acquired at low cost with consumer-grade devices such as smartphones and panoramic cameras (\eg{} Insta360 X5). These captures are fed into the \abotgs{} pipeline, which can stably reconstruct over $10^5$~m$^2$ of contiguous indoor or outdoor space per project. The resulting scenes form the backbone of our indoor and street-level training pool.

\paragraph{Proprietary urban / aerial 3DGS scenes}
For aerial and city-block coverage, we leverage AMAP's in-house aerial 3DGS assets, reconstructed from large-scale aerial photography by \abotgs{}. These aerial 3DGS tiles provide citywide and aerial scenery that purely ground-level captures cannot reach, allowing \method{} to learn long-range outdoor exploration consistent with real-world cities. We additionally incorporate satellite-derived city models~\cite{fromorbit2ground2025} as a supplementary source of coarse urban geometry.

\paragraph{Public 3DGS / panorama datasets}
We further include open-source panoramic datasets~\cite{chang2017matterport3dlearningrgbddata,dai2017scannet} and 3DGS datasets~\cite{InteriorGS2025}, which mainly cover indoor rooms and complement our proprietary assets with additional real-world layout and appearance variation. All assets undergo unified coordinate normalization, scale alignment, and metadata cleansing before entering the rendering pipeline.

\paragraph{Bootstrapped stylistic 3DGS scenes}
Finally, we crawl licence-permissive internet panoramas, drone clips, and street-view sequences. After filtering and deduplication, these sparse inputs are lifted into 3DGS scenes by the \method{} pipeline itself; the same data pipeline then selects high-quality outputs and feeds them back into the training set, extending the long tail of styles and diversity (fantasy, cartoon, historical reconstructions, and other film-style scenes).

\subsection{Training-Sample Rendering from 3D Assets}
 \label{sec:data_sample_rendering}

Given the curated 3DGS asset bank, we render 81-frame panoramic videos along sampled camera trajectories. For each target video, we construct the same conditioning interface used by the panoramic video generator in \cref{sec:method_video}: trajectory-aligned controls and scene memory. Each training sample therefore comprises a target video, camera poses along the trajectory, point-cloud RGB renderings, depth maps, point-cloud visibility masks, and a small set of posed reference panoramas used as memory.

\paragraph{Scene-normalized camera trajectories}
We apply the geometry-grounded planner introduced in \cref{sec:method_traj} to generate collision-free trajectories within each scene. Since the 3DGS assets represent static scenes, all temporal variation in the rendered videos is induced by camera motion. We adopt translation-only trajectories and keep the panoramic camera orientation fixed throughout each clip. Unlike perspective cameras, whose rotations change the visible field of view, a panoramic camera already observes the complete viewing sphere. Fixing its orientation therefore avoids unnecessary spherical image rotation while preserving the parallax, occlusion, and disocclusion induced by camera translation.

To cover different camera-motion magnitudes under the fixed 81-frame duration, we estimate the scene scale from depth and define a scene-adaptive unit translation. We then sample the traversal rate from \(\{1\times,2\times,3\times,4\times\}\). The point cloud and camera translation share the same scene scale; hence, a global rescaling of depth induces a corresponding rescaling of the trajectory without changing the relative reprojection geometry. The same scale-normalized trajectory parameterization is applied during training and inference. The camera poses along the trajectory are stored and later converted into camera-pose ray maps for the video generator.

\paragraph{Trajectory-aligned control rendering}
The target RGB video is rendered directly from the 3DGS asset along the sampled trajectory. To construct dense controls, we render a reference RGB panorama and its aligned depth from the same asset, lift the reference RGB-D panorama into a colored point cloud, and reproject it to each target camera pose using a radial z-buffer. This produces trajectory-aligned point-cloud renderings, depth maps, and point-cloud visibility masks. These streams share the same projection and visibility computation, ensuring consistent occlusion across conditioning signals.

Because the controls are derived from a reprojected reference-view point cloud, they retain the partial-observation characteristics encountered at inference time. Camera translation exposes regions absent from the point cloud, while some source-visible points may become invalid in the target view. The point-cloud visibility mask therefore indicates projected point-cloud support rather than guaranteed correctness. The generator is trained to complete missing regions and refine imperfect reprojections instead of treating the rendered controls as immutable targets.

\paragraph{Scene-memory sampling}
To train the reference-memory pathway, we additionally sample up to four posed reference panoramas from nearby viewpoints in the same 3DGS scene. Since each panorama observes the full \(360^\circ\) surrounding field, camera-center proximity provides a natural criterion for selecting reference observations with strong spatial overlap. Each reference panorama is paired with its camera pose and used as a scene-memory condition. This simulates the inference-time setting in which previously generated panoramas are stored and retrieved to preserve appearance, layout, and object identity during long-range exploration.

\subsection{Quality Control}
\label{sec:data_quality}

Even with high-quality assets, careless rendering produces frames that hurt downstream training (occluded by floaters, geometrically out-of-bounds, or texture-poor). We deploy a two-tier filtering pipeline. (i) Geometric coverage filtering removes frames with insufficient valid depth or trajectories that intersect reconstructed geometry. (ii) VLM-based perceptual scoring runs a Vision--Language Model on each candidate clip to score texture sharpness, artifact absence, and aesthetic quality; clips below the threshold are recycled with adjusted trajectory parameters. Final corpora are dataset-balanced across scene category (indoor / outdoor street / aerial / scenic / stylized) to prevent any one regime from dominating the generative prior.

\subsection{One Data Engine, Multiple Modules}
\label{sec:data_cross_module}

A distinctive feature of this data engine is that all supervision is fully rendered from 3D data, providing aligned training signal to multiple learnable modules in \method{}. Specifically, the same (RGB, point-map, trajectory) tuples are consumed by: (i) the DiT-based panorama generator (\cref{sec:method_pano}), which learns text/image-conditioned equirectangular synthesis; (ii) the 14B panoramic video diffusion model (\cref{sec:method_video}), conditioned on point-cloud renderings that are, by construction, pixel-aligned with RGB targets; and (iii) the 3DRL post-training stage (\cref{sec:method_3drl}), which fine-tunes the video generator on the same rendered panoramic corpus with a reconstruction-free geometry-consistency reward. Because every module observes training signal derived from 3DGS assets rather than 2D imagery, the 3D consistency of each module is strongly reinforced, and end-to-end 3DGS generation quality is improved as a result.

\section{Method}
\label{sec:method}

This section details the algorithmic core of \method{}. \cref{fig:pipeline_overview} sketches the full pipeline, which we organize into seven tightly integrated stages: (\cref{sec:method_input}) defining the Spatial Generative Primitive and the two input pathways that produce it, (\cref{sec:method_pano}) generating high-fidelity $360^\circ$ panoramas from text or perspective images, (\cref{sec:method_traj}) designing exploration trajectories grounded in scene geometry, (\cref{sec:method_video}) generating controllable, 3D-consistent panoramic exploration videos, (\cref{sec:method_3drl}) post-training the generator with 3D Reinforcement Learning, (\cref{sec:method_gs}) reconstructing a clean 3DGS world from the generated panoramic video with our \abotgs{} repair-and-refine engine, and finally (\cref{sec:method_physics}) extracting physical attributes and rendering effects (occupancy collision mesh, portal teleportation) that turn the world into an interactive product.

\subsection{The Spatial Generative Primitive}
\label{sec:method_input}

\paragraph{Definition}
We define the \spatialprim{} as the tuple
\begin{equation}
\mathcal{S} \triangleq \big(\mathbf{P}, \mathbf{G}\big),
\end{equation}
where $\mathbf{P}$ is a high-quality panorama and $\mathbf{G}$ is a spatial point cloud that encodes the scene's metric geometry. The panorama supplies appearance over the full $360^\circ$ field-of-view in a single image; the point cloud supplies dense 3D structure. This factorization delivers a near-optimal compression of a 3D space for downstream generative use, mirroring the consumer-facing real-estate viewing experience where panoramic stills plus sparse geometry already deliver a compelling sense of place. Each SGP is additionally associated with the camera pose $\mathbf{T}$ from which $\mathbf{P}$ was captured; the pose is not part of the generative primitive itself but serves as the spatial anchor that enables metric registration and composition across multiple SGPs. Trajectories are not part of the primitive itself: they are produced on top of $\mathcal{S}$ by the geometry-grounded planner of \cref{sec:method_traj} and consumed by the video generator of \cref{sec:method_video} as a separate exploration plan $\boldsymbol{\tau} = \{(\mathbf{R}_t, \mathbf{t}_t)\}_{t=1}^{T}$.

\paragraph{Composability}
Multiple SGPs compose into a larger world $\mathcal{W} = \{\mathcal{S}_1, \ldots, \mathcal{S}_K\}$ through spatial adjacency within a shared metric frame. Because every SGP carries metrically consistent geometry, neighboring primitives can be registered and stitched to extend scene coverage (\eg{} from a single room to a full multi-room interior, or from one street segment to an entire city block). This compositional design means \method{} handles both single-room experiences (a single SGP) and city-scale walkthroughs (a graph of SGPs) under the same representation, making the SGP a true compositional primitive for 3D world generation.

\paragraph{Minimal-input path: text or single image $\rightarrow$ one SGP}
For a text prompt or a single perspective image, we first synthesize a coherent $360^\circ$ panorama $\mathbf{P}$ via the panorama generator described in \cref{sec:method_pano}. Given $\mathbf{P}$, we estimate the dense point map $\mathbf{G}$ with a MoGe-2-style monocular geometry estimator~\cite{moge2}: a transformer-based depth predictor consumes the equirectangular panorama in cube-map projection, predicts per-face depth, and re-projects back to a unified panoramic point map. We fine-tune the predictor on a synthetic single-panorama-depth corpus, where ground-truth panoramic depth is read directly from \abotgs{}-reconstructed scenes. The resulting $(\mathbf{P}, \mathbf{G})$ pair constitutes one complete SGP. This path dramatically lowers the input barrier: a single photograph---or even a text description---suffices to initialize a navigable 3D world. It is especially valuable for scenes that cannot be physically captured today, such as historical landmarks, fictional environments, or scenic viewpoints accessible only in imagination.

\paragraph{Rich-input path: multi-image or video $\rightarrow$ multiple SGPs}
For richer inputs---multi-view photo sets or short smartphone videos (\eg{} shop owners self-capturing their store interiors for AMAP)---we employ a sparse-to-dense reconstruction pipeline powered by \abotgs{}. Incoming frames are first key-frame-sampled and jointly registered into a metric coordinate frame, yielding per-frame camera poses and a sparse point cloud. \abotgs{} then performs a fast sparse-to-dense 3DGS reconstruction, producing a partial but geometrically faithful scene. From this reconstruction we render panoramas at multiple representative viewpoints (POIs); regions not covered by the original observations are completed via guided inpainting. Each rendered panorama together with its corresponding point map forms one SGP. The collection of SGPs shares a common metric frame and composes spatially into a multi-room world $\mathcal{W}$ that faithfully preserves the photometric and geometric evidence from the original capture. This path prioritises fidelity: because every SGP is anchored in real observations, the resulting world is a serious reconstruction rather than a hallucinated guess.

\subsection{Panorama Generation from Perspective Input}
\label{sec:method_pano}

A $360^\circ$ equirectangular panorama captures the full field-of-view from a single viewpoint, preserving global spatial context and semantic relationships that a narrow perspective image cannot convey. This holistic representation serves as the appearance backbone of the SGP and provides the spatial consistency required for coherent trajectory planning and video synthesis downstream. We propose ABot-Pano, which synthesizes high-fidelity panoramas from text prompts or single perspective photographs.

\textbf{ABot-Pano} is a multimodal controllable panoramic image generation framework built upon the Diffusion Transformer (DiT)~\cite{peebles2023dit} architecture. Departing from prior approaches that rely exclusively on textual or single visual conditions, ABot-Pano achieves unified generation of high-fidelity equirectangular projection (ERP) panoramas from text, perspective images, or their combinations, via an adaptive condition injection mechanism and a dual-modal independent decoupling training strategy.

\paragraph{Adaptive Condition Injection Mechanism}
To ensure precise adherence to perspective conditions, we devise a Zero-Convolution-based residual injection scheme. The input perspective image is first geometrically projected into ERP format and encoded by a VAE; the Control branch then generates layer-wise residual signals, which are injected into the backbone network via scalable additive fusion. When the conditional input is an empty image, the model gracefully degrades to a pure text-to-panorama or unconditional generation mode without requiring architectural modifications or weight reloading. This design enables a single model to span the continuous spectrum from precise viewpoint control to open-ended semantic generation.

\paragraph{Dual-Modal Independent Decoupling Training Strategy}
To accommodate flexible multimodal input combinations, we apply independent stochastic dropout to textual and perspective conditions during training:
\begin{itemize}[nosep]
  \item Caption Dropout compels the model to learn pure visual extrapolation.
  \item Image Dropout preserves pure textual generation capability.
  \item Joint retention optimizes the synergy between cross-modal semantic alignment and spatial constraints.
\end{itemize}
This strategy establishes ABot-Pano as a unified multimodal generative model that maintains semantic consistency and generation quality regardless of whether text, perspective images, or both are provided at inference time.

\paragraph{Panoramic Geometric Regularization}
To address the inherent artifacts of ERP representations, we incorporate horizontal circular padding alongside two geometry-aware losses—Distortion-aware Cube Loss and Rotation-consistent Yaw Loss~\cite{dit360,feng2023diffusion360}—during training. These components enable the model to natively perceive 360° boundary continuity while enhancing distortion correctness in polar regions and global rotational equivariance.

\paragraph{Training data}
We curate a hybrid training corpus combining (i)~high-resolution real-world panoramas with rigorous quality filtering to remove stitching artifacts and visible capture equipment, and (ii)~synthetic panoramas rendered from diverse 3D environments to broaden semantic coverage. This dual-source strategy is consistent with emerging community practice while adding AMAP-specific map-linked scene categories that are scarce in public datasets.

\subsection{Exploration Trajectory Design}
\label{sec:method_traj}

While a single equirectangular panorama captures the comprehensive $360^\circ$ visual representation of a scene, it remains devoid of intrinsic traversability cues. Advancing from passive observation to active exploration requires robust spatial reasoning that transcends simple camera-path sampling. We therefore cast trajectory planning as an agentic reasoning problem. Conditioned on a single static observation, the agent reconstructs the scene geometry, isolates semantically salient regions, builds a navigability graph, and synthesizes a sequence of exploration trajectories optimized to simultaneously maximize spatial coverage, exploration efficiency, and reconstruction quality.

\begin{figure}[t]
    \centering
    \includegraphics[width=0.96\textwidth]{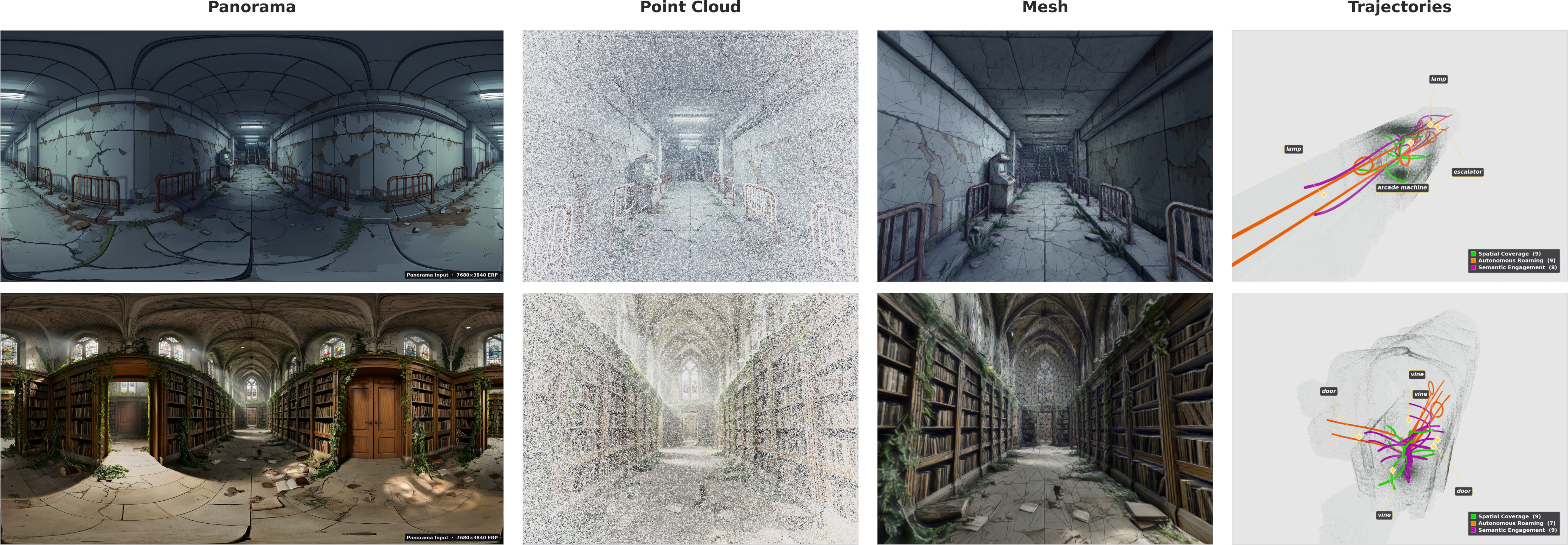}
    \caption{Panoramic agentic trajectory planning. Starting from a single panoramic image (left), we estimate a dense point cloud using MoGe-2~\cite{moge2}. This geometric representation is then converted into a navigable 3D mesh through RecastNavigation~\cite{mononen2009recast}. Finally, diverse agent trajectories are planned within this reconstructed environment to satisfy specific exploration goals.}
    \label{fig:trajectoryplanning}
\end{figure}

\paragraph{Geometry as the agent's primary language}
Prior to any semantic reasoning, the agent recovers a metrically consistent geometric representation of the scene. Specifically, the input panorama is processed by MoGe-2~\cite{moge2}, a metric depth foundation model, which predicts a dense distance field together with per-pixel surface normals over the entire sphere. In contrast to purely semantic cues, this estimate yields a metrically grounded characterization of scale, shape, and spatial enclosure—properties that are indispensable for downstream physical reasoning but inaccessible from appearance alone. Building on the predicted distance field, the agent back-projects the equirectangular $2{:}1$ pixel grid into a unified global point cloud and a corresponding triangulated mesh, thus promoting the two-dimensional panoramic observation to an explicit three-dimensional representation that supports geometry-aware inference in subsequent stages.

\paragraph{From geometry to affordance}
Recovering the geometry of the scene is necessary but insufficient for planning the motion. The agent must further extract navigability: which surfaces are walkable, which regions are obstructed, and how reachable locations connect. We process the reconstructed mesh through RecastNavigation~\cite{mononen2009recast} to obtain a navigation mesh---a graph of walkable polygons---then densely sample it into a spatial graph $\mathcal{G}$ on which shortest-path queries become tractable. At this stage, the world model transitions from a passive reconstruction to an actionable spatial structure over which the agent can reason about routes, distances, and reachability.

\paragraph{Semantic curiosity}
An agent that solely enlarges spatial coverage produces geometrically thorough but semantically undifferentiated trajectories. To introduce intentionality, we equip the agent with a semantic perception module: it queries a vision-language model (Qwen3-VL-8B)~\cite{Qwen3-VL} to identify salient objects in the panorama. Each named entity is detected and localized in 3D space through a detection-segmentation pipeline (Grounding DINO~\cite{liu2023grounding} $\to$ SAM3~\cite{carion2025sam3segmentconcepts} $\to$ depth back-projection) as shown in \cref{fig:trajectoryplanning} and registered as a goal in the navigation graph. The agent thus acquires both a navigable map and a set of semantically grounded destinations, the prerequisites for purposeful exploration.

\paragraph{Planning as a three-way trade-off}
With the navigable graph and semantic goals in place (\cref{fig:trajectoryplanning}), the agent resolves three competing objectives through trajectory design. Spatial coverage, visualized as green trajectories, ensures that no significant navigable region remains unobserved; this is enforced through farthest-point sampling and bearing-sector partitioning on $\mathcal{G}$. Semantic engagement, depicted by purple trajectories, directs the agent to approach discovered objects at informative distances and viewing angles. Reconstruction quality, represented by orange trajectories (corresponding to autonomous roaming), mandates that the planned paths yield a sequence of observations with sufficient parallax and view overlap, thereby providing the dense multi-view supervision necessary for robust 3DGS optimization. 

To achieve this trade-off, the agent generates a diverse pool of candidate trajectories and evaluates them against the joint objective. It subsequently filters the candidates to guarantee collision safety via KD-tree proximity queries against the point cloud, while enforcing mutual diversity through cosine-based deduplication over the full pose sequences. Ultimately, the resulting trajectories are dynamically synthesized to accommodate the unique geometry and semantics of the scene, rather than being selected from a fixed library of motion primitives.

\paragraph{Trajectory refinement and robustness}
The coarse sequence of discrete camera poses is smoothed into a continuous trajectory through cubic B-spline fitting and subsequently resampled into 21 equidistant keyframes. Gaze direction is determined by trajectory intent: coverage-oriented paths look ahead along their tangent; object-directed paths lock onto the target centroid. Physical plausibility constraints (bounded inter-frame rotation, bounded pitch, minimum arc length) eliminate degenerate candidates, and a final diversity pass retains the most mutually complementary subset. When the NavMesh graph is too small for path planning the agent falls back to local viewpoint variation; when semantic targets lie beyond reachable range it redirects effort toward coverage---ensuring a useful output under degraded input conditions.

\paragraph{Comparison with related schemes}
Compared with the WorldNav planner of HY-World 2.0~\cite{hyworld22026} that drives a perspective generator over long sequences, our panoramic trajectory module trades sequence length for field of view: each step covers more space, the total exploration is shorter, and the resulting computational footprint is markedly lower.

\subsection{Panoramic Exploration Video Generation}
\label{sec:method_video}

We now turn to the centerpiece of \method{}: the panoramic video generator that transforms an SGP into a high-quality exploration video. Given the SGP point cloud \(\mathbf{G}\) and a planned exploration trajectory \(\boldsymbol{\tau}\) produced by the geometry-grounded planner of \cref{sec:method_traj}, the generator synthesizes an equirectangular (ERP) panoramic video \(\hat{V}\in\mathbb{R}^{T\times H\times W\times3}\) that is both visually rich and tightly aligned with the underlying 3D geometry.

A key design decision is to use panoramic video as the exploration carrier, an approach also taken by recent panoramic world models~\cite{omniroam2025,xia2025panowan,wu2026360anything}, rather than generating and fusing multiple narrow-FOV perspective views. Each panoramic frame captures the full \(360^\circ\) surrounding field from a single camera center, providing higher spatial coverage per generated frame. This reduces viewpoint blind spots caused by narrow-FOV sampling and preserves native spatio-temporal continuity by representing the surrounding view within a single coherent video stream. The resulting omnidirectional observations also provide dense cross-view overlap, improving exploration efficiency and simplifying downstream 3DGS reconstruction (\cref{sec:method_gs}). At the same time, panoramic video generation must address two practical challenges: maintaining 3D consistency in a model that is not inherently 3D-aware, and recovering high-resolution details from ERP frames constrained by memory and compute. In this section, we first describe the base panoramic video generator and its geometry-aware conditioning design. The two remaining challenges are then addressed in subsequent sections through 3DRL post-training for geometric consistency (\cref{sec:method_3drl}) and 3DGS repair for detail enhancement and residual artifact removal (\cref{sec:method_gs}).

\begin{figure}[t]
    \centering
    \includegraphics[width=0.96\textwidth]{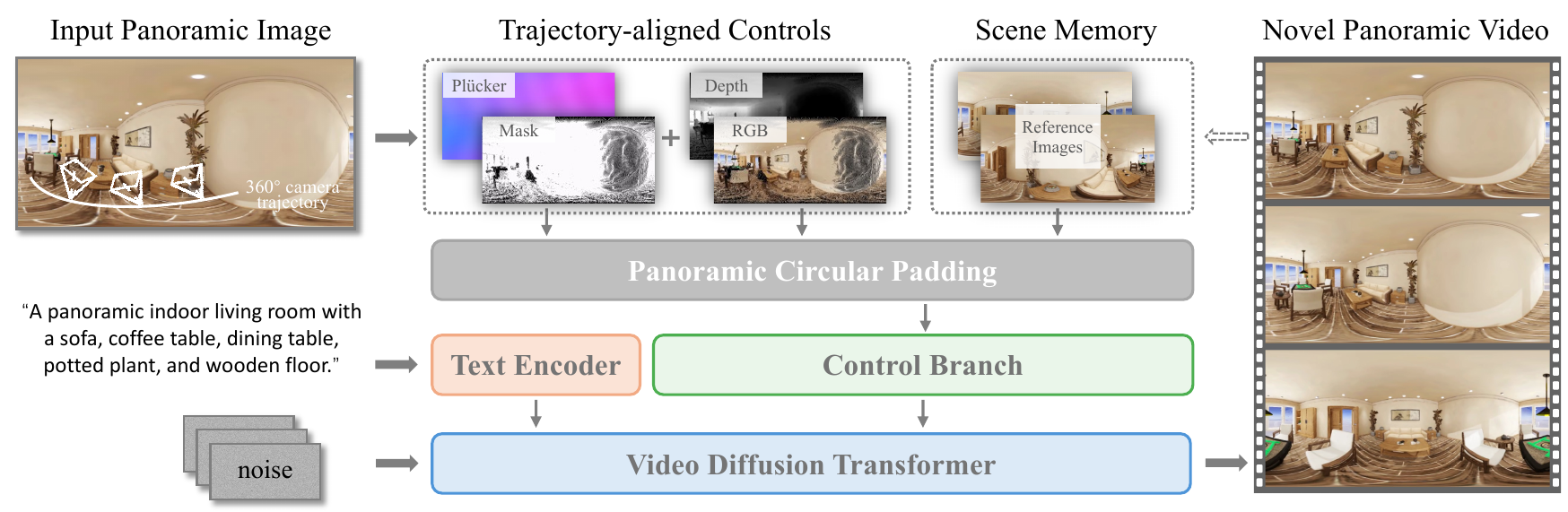}
    \caption{Panoramic video generation with trajectory-aligned controls and scene memory. Given an input panoramic image and a planned camera trajectory, we construct trajectory-aligned controls, including point-cloud RGB renderings, depth maps, point-cloud visibility masks, and camera-pose ray maps. Nearby reference panoramas from generated viewpoints are retrieved as scene memory to provide appearance and layout anchors. These signals are processed with panoramic circular padding and integrated by the control branch to guide the video diffusion transformer, producing a 3D-consistent novel panoramic video.}
    \label{fig:video_pipeline}
\end{figure}

\subsubsection{Backbone and Conditioning}
\label{sec:backbone_cond}

\paragraph{Backbone}
We build on the VACE~\cite{vace2025,wan2025wan} family of controllable latent video diffusion models~\cite{rombach2022ldm} and adopt a 14B-parameter base model. VACE provides a strong controllable generation prior and a control-branch architecture that can be initialized for structured video conditioning. We extend this control pathway to panoramic video generation by integrating trajectory-aligned controls and scene memory, enabling geometry-aware and reference-guided conditioning while keeping the main diffusion backbone unchanged.

\paragraph{Trajectory-aligned controls}
For each target camera pose along the planned trajectory \(\boldsymbol{\tau}\), we construct a set of spatially aligned control signals that provide dense frame-wise appearance, geometry, visibility, and camera-pose guidance.

\begin{itemize}
    \item \textbf{Point-cloud rendering.}
    Inspired by point-conditioned video diffusion methods such as ViewCrafter~\cite{yu2024viewcrafter}, GEN3C~\cite{ren2025gen3c}, and Uni3C~\cite{cao2025uni3c}, we render the SGP point cloud \(\mathbf{G}\) from each target panoramic camera pose along \(\boldsymbol{\tau}\), yielding an RGB control video \(\{R_t\}_{t=1}^{T}\). This trajectory-aligned rendering provides coarse appearance guidance grounded in observed scene evidence: it anchors well-observed regions while leaving the generator free to refine imperfect reprojections and synthesize disoccluded or previously unseen content.

    \item \textbf{Depth video.}
    A per-frame depth map \(\{D_t\}_{t=1}^{T}\) is rendered from the same trajectory. This stream supplies complementary geometric structure, helping the generator reason about scene layout, occlusion boundaries, and depth-aware temporal consistency.

    \item \textbf{Point-cloud visibility mask.}
    Following AnyRecon~\cite{chen2026anyrecon}, we use a binary mask video \(\{K_t\}_{t=1}^{T}\) to indicate whether each ERP pixel is supported by projected point-cloud evidence. Concatenated with the rendered point-cloud observations, this mask provides a spatially aligned support signal: unsupported regions require generative completion, while supported regions offer useful but imperfect appearance and geometry hints due to occlusion changes, sparse geometry, depth noise, or reprojection distortion. Thus, the mask guides the model without acting as a hard editing constraint.

    \item \textbf{Camera-pose ray map.} 
    Inspired by the panoramic \Plucker{} embedding of CamPVG~\cite{ji2025campvg}, we encode each target camera pose as a dense per-pixel ray map under ERP projection. Each ERP pixel is mapped to a unit viewing direction on the panoramic sphere according to the camera-coordinate convention of our rendering pipeline. Given the target camera pose, the ray direction and origin are transformed into world space, and the corresponding ray is represented by its \Plucker{} coordinates:
    \begin{equation}
    p_t(u,v)=
    \left[
    d_{w,t}(u,v),\ 
    o_{w,t} \times d_{w,t}(u,v)
    \right]\in\mathbb{R}^{6},
    \qquad
    C = \{p_t(u,v)\}_{t,u,v}
    \in \mathbb{R}^{T\times H\times W\times6}.
    \end{equation}
    Here, \(o_{w,t}\) denotes the world-space camera center at frame \(t\). Compared with raw extrinsics or global pose tokens, this representation explicitly associates every panoramic pixel with its 3D viewing ray, providing trajectory-aware geometric guidance to the video generator.
\end{itemize}

\paragraph{Scene memory}
In addition to trajectory-aligned controls, we maintain a small set of posed reference images as scene memory. During long-range panoramic exploration, previously generated panoramic images along the trajectory are stored together with their camera poses. For each new generation segment, we retrieve up to four nearby reference observations:
\begin{equation}
\mathcal{M}
=
\{(I_j^{\mathrm{ref}}, T_j^{\mathrm{ref}})\}_{j=1}^{N_{\mathrm{ref}}},
\qquad
N_{\mathrm{ref}}\leq4.
\end{equation}
Since each panoramic image observes the full \(360^\circ\) surrounding field, camera-center proximity provides a natural criterion for selecting reference observations with strong spatial overlap. These posed reference images act as memory anchors rather than regeneration targets, helping the model preserve appearance, object identity, texture, and layout when synthesizing novel views. Their associated camera poses ground the remembered visual evidence in explicit spatial locations, reducing drift and improving multi-view consistency across the exploration trajectory.

\paragraph{Control-branch integration}
We initialize a panoramic control branch from the VACE control weights and fine-tune it for geometry-aware panoramic video generation. The RGB rendering \(\{R_t\}\) and depth video \(\{D_t\}\) are VAE-encoded as latent control features, while the point-cloud visibility mask \(\{K_t\}\) and camera-pose ray embedding \(C\) are downsampled into a compact mask-camera embedding. The retrieved scene memory \(\mathcal{M}\) is encoded as reference-memory conditions, grounding nearby visual evidence with its camera poses. These features are fused by the control branch and injected into the video diffusion transformer to guide trajectory-aware, scene-consistent panoramic video synthesis.

\paragraph{LightX2V acceleration}
Following the success of step-distilled diffusion video models, we apply LightX2V-style consistency distillation~\cite{lightx2v2025} to compress the panoramic video sampler from 50 steps down to 4 steps. Together with moderate spatial and temporal patch reduction, this brings inference cost into the consumer-GPU regime while keeping perceptual quality essentially unchanged. On a 4$\times$4090 node, the model is kept resident in VRAM to reduce loading overhead, and inference takes about 12 minutes per scene.

\subsubsection{Seamless Wrap-around}
\label{sec:circular_padding}

The conditioning design above grounds the generated video in 3D geometry and scene memory. However, panoramic generation also requires respecting the topology of the ERP image itself. The ERP projection imposes a circular boundary constraint: the left and right image borders are physically adjacent on the viewing sphere, yet a flat image representation treats them as disconnected boundaries. Naively processing a panorama with standard convolutional layers, which typically use zero-padding at image boundaries, introduces discontinuities in the latent representation and can lead to visible vertical seam artifacts at the \(\phi=\pm180^\circ\) longitude. Prior panoramic image and video generation works have widely observed this wrap-around discontinuity and introduced circular blending, latent rotation, rotated denoising, or padded decoding to improve boundary consistency~\cite{feng2023diffusion360,wang2024360dvd,xia2025panowan,wu2026360anything}. These analyses suggest that boundary artifacts can arise not only in pixel space, but also in latent encoding and denoising, where standard padding and flat-image processing break the circular topology of ERP panoramas.

We therefore enforce circular continuity at three stages of the generation pipeline: VAE encoding, latent denoising, and VAE decoding.

\paragraph{Circular padding at VAE encoding}
During training, we apply circular padding along the width (longitude) dimension before VAE encoding. Specifically, the rightmost \(W/8\) columns are prepended to the left boundary and the leftmost \(W/8\) columns are appended to the right boundary, where \(W\) is the original ERP width. After encoding, the latent columns corresponding to the padded regions are cropped to recover the original latent width. This padding-encode-crop procedure allows the VAE encoder's convolutional kernels to observe continuous panoramic context across the wrap-around boundary, producing seam-free latent representations. We apply the same procedure to panoramic conditioning inputs, including point-cloud renderings, depth maps, visibility masks, camera-pose ray maps, and reference-memory panoramas.

\paragraph{Latent uniform shift during denoising}
During inference, at each denoising step, the latent tensor and its conditioning context are jointly rolled along the width dimension by a constant per-step offset of \(\lfloor W_{\ell}/N \rfloor\) columns, where \(W_{\ell}\) is the latent width and \(N\) is the total number of denoising steps. The cumulative shift is undone after denoising to restore the original alignment. By exposing different longitudes to the latent boundary over the denoising process, this strategy prevents seam-related errors from accumulating at a fixed location and suppresses semantic inconsistencies at the wrap-around boundary.

\paragraph{Circular padding at VAE decoding}
Before decoding the denoised latent, we apply circular padding in latent space: the rightmost \(W_{\ell}/8\) columns are prepended to the left boundary and the leftmost \(W_{\ell}/8\) columns are appended to the right boundary, where \(W_{\ell}\) denotes the latent width. The padded latent is decoded by the VAE, and the corresponding pixel-space padding is cropped from the output. This allows the VAE decoder's convolutional kernels to access continuous context across the wrap-around boundary, suppressing pixel-level seam artifacts.

Together, these mechanisms enforce circular continuity throughout latent encoding, diffusion denoising, and latent decoding, yielding seamless panoramic videos without post-hoc blending.
\subsection{3D Reinforcement Learning Post-Training}
\label{sec:method_3drl}

Standard flow-matching training improves per-frame realism but does not explicitly enforce temporal or geometric consistency. Thus, panoramic videos may appear plausible frame by frame while still exhibiting flickering, inconsistent motion, or geometry hallucinations. Reconstruction-based rewards provide a direct alternative, but they are computationally expensive and can favor blurred or over-smoothed videos that are easier to fit.

We introduce 3D Reinforcement Learning (3DRL), a post-training stage that optimizes a reconstruction-free geometry-consistency reward based on forward-backward optical-flow cycle consistency. A geometrically coherent video should allow pixels to be tracked forward and then back to their original locations. Optimizing this reward encourages temporally stable and 3D-aware panoramic videos without relying on explicit 3D reconstruction.

\paragraph{Reward design}
We adopt DiffusionNFT~\cite{black2023ddpo,diffusionnft2025}-style policy optimization with a composite 3D-aware reward. Given a generated panoramic video $\hat{V}$, our goal is to evaluate whether its temporal motion is consistent with a coherent underlying 3D scene, without fitting an explicit 3D representation. To this end, we use forward-backward optical-flow cycle consistency as a proxy for geometric coherence. Our reward consists of two complementary terms: a cycle-consistency reward for reconstruction-free geometric coherence, and a fidelity reward to prevent static-video reward hacking and distributional drift.

\begin{itemize}
    \item \textbf{Cycle-consistency reward.}
    Directly estimating optical flow on ERP panoramas is unreliable due to projection distortion and horizontal boundary wrapping. We therefore convert the generated panorama video $\hat{V}$ into $K$ perspective videos by uniformly sampling pinhole views along the equator, denoted as $\{\hat{V}^k\}_{k=1}^{K}$. For each perspective video $\hat{V}^k$, we sample $N$ frame pairs $(t_i, t_i+s)$ with temporal stride $s$. Given a sampled pair of frames $I_{t_i}^k$ and $I_{t_i+s}^k$, we estimate forward and backward optical flows using RAFT~\cite{teed2020raft}:
    \begin{equation}
    F_{t_i \rightarrow t_i+s}^{k}
    =
    \mathrm{RAFT}\!\left(I_{t_i}^{k}, I_{t_i+s}^{k}\right),
    \qquad
    F_{t_i+s \rightarrow t_i}^{k}
    =
    \mathrm{RAFT}\!\left(I_{t_i+s}^{k}, I_{t_i}^{k}\right).
    \end{equation}

    The forward-backward cycle error at pixel $x$ is:
    \begin{equation}
    e_i^k(x)
    =
    \left\|
    F_{t_i \rightarrow t_i+s}^{k}(x)
    +
    \mathcal{W}\!\left(
    F_{t_i+s \rightarrow t_i}^{k},
    F_{t_i \rightarrow t_i+s}^{k}
    \right)(x)
    \right\|_2 ,
    \end{equation}
    where $\mathcal{W}(F,G)(x)=F(x+G(x))$ denotes bilinear sampling of flow field $F$ at the location displaced by $G(x)$. Pixels whose warped coordinates fall outside the image boundary are excluded from the valid set $\Omega_i^k$. We average the cycle error over valid pixels, sampled frame pairs, and perspective views:
    \begin{equation}
    \bar{e}
    =
    \frac{1}{K}
    \sum_{k=1}^{K}
    \frac{1}{N}
    \sum_{i=1}^{N}
    \frac{1}{|\Omega_i^k|}
    \sum_{x \in \Omega_i^k}
    e_i^k(x).
    \end{equation}

    The cycle-consistency reward is then normalized as:
    \begin{equation}
    r_{\mathrm{cycle}}
    =
    \mathrm{clip}\!\left(
    1 - \frac{\bar{e}}{\tau_{\mathrm{cycle}}},
    0,
    1
    \right).
    \end{equation}

    \item \textbf{Fidelity reward.}
    Optimizing only the cycle-consistency reward may lead to reward hacking, since nearly static videos can produce small optical-flow cycle errors without improving 3D coherence. To prevent this degenerate solution and preserve the generation distribution of the pretrained model, we compare the RL-generated video $\hat{V}$ with the reference video $\hat{V}_{\mathrm{ref}}$ produced by the frozen pre-RL model using the LPIPS perceptual metric~\cite{zhang2018lpips}:
    \begin{equation}
    r_{\mathrm{fid}}
    =
    1 - \mathrm{LPIPS}(\hat{V}, \hat{V}_{\mathrm{ref}}).
    \end{equation}
\end{itemize}

Together, these two terms encourage temporally and geometrically coherent videos while preserving the generation distribution of the pretrained model.

\subsection{Panoramic-Video to 3DGS Reconstruction}
\label{sec:method_gs}

The generated panoramic exploration video is finally lifted into a clean 3DGS world by our \abotgs{} engine. The reconstruction pipeline comprises three stages: (i)~recovery of the camera trajectory and a sparse seed point cloud directly from the panoramic video, initializing a vanilla 3DGS; (ii)~an iterative FLUX~\cite{flux2024}-based repair-and-refit loop leverages deterministic single-step diffusion to super-resolve texture detail and suppress inconsistencies inherited from the generative video; and (iii)~a suite of wild-scene robustness modules ensures that the same engine operates reliably on both synthetic panoramic videos and real-world Insta360 captures, without scene-specific tuning. This universality---from generative content to large-scale professional captures---is a key differentiator of \abotgs{}.

\paragraph{From panoramic video to coarse 3DGS}
We first recover the camera trajectory and a sparse seed point cloud from the panoramic video $\hat{V}$ via structure from motion~\cite{schoenberger2016colmap}; the planned trajectory $\boldsymbol{\tau}$ from \cref{sec:method_traj} provides a strong prior that we leverage for fast convergence. The seed point cloud then initializes a vanilla 3DGS~\cite{3DGS} optimization, producing a coarse reconstruction in a few minutes per scene. In contrast to feed-forward reconstruction models that regress scene geometry in a single pass~\cite{liu2025worldmirror,zhang2026worldstereo}, this per-scene optimization maximizes fidelity to the generated video and scales to large real-world captures.

\paragraph{Iterative FLUX-based Gaussian repair}
Generative panoramic videos suffer from low effective spatial resolution and imperfect multi-trajectory consistency. Since each ERP frame distributes its pixels over the full $360^\circ$ field of view, local perspective crops contain limited high-frequency detail. In addition, independently generated trajectories often exhibit photometric and geometric discrepancies, which destabilize vanilla 3DGS optimization. We therefore adopt an iterative repair-and-refit strategy inspired by Difix3D+~\cite{difix3dplus2025}.

At each iteration, a FLUX-based refiner takes as input a training view rendered from the current 3DGS and the corresponding high-quality reference crop from the original panorama, refining it to serve as guidance for the next iteration. Both images are encoded by a shared VAE and processed jointly by a shared diffusion transformer. Instead of denoising from random noise, the refiner starts from the low-quality latent and deterministically produces a repaired high-resolution view in a single forward pass. To improve robustness in occluded and weakly textured regions, the refiner is trained with random patch dropout, encouraging it to recover missing structures from contextual and geometric cues. The repaired views are then used as additional supervision to re-optimize the 3DGS. We perform two repair-and-refit passes: the first pass mainly removes severe artifacts and lifts the effective resolution, while the second pass further reduces residual photometric/geometric drift using the improved intermediate 3DGS as a more stable canvas. This iterative scheme yields sharper textures, fewer floaters, and stronger cross-view consistency than single-pass refinement.

\paragraph{Wild-scene robustness from \abotgs{}}
Beyond iterative repair, \abotgs{} includes several robustness modules that allow the same reconstruction pipeline---camera recovery, view repair, and Gaussian optimization---to operate on both generated panoramic videos and real-world Insta360 captures without architectural changes.

\begin{itemize}
    \item \textbf{Sky modelling.}
    Sky regions occupy a large portion of panoramic images but provide weak depth and multi-view constraints. Directly modeling them with standard 3D Gaussians often leads to translucent floaters and depth ambiguity. We separate sky and foreground using masks from depth-anything3~\cite{da3depth2025}. Explicit sky Gaussians are constrained to a spherical-shell distribution with tangent alignment, while implicit sky representations are also supported. During optimization, sky Gaussians are gradually pushed toward transparency and pruned, after which the foreground is refined in a merged pass. This decouples foreground geometry from sky appearance and enables independent sky editing.

    \item \textbf{Appearance calibration.}
    Both generated and real panoramic videos suffer from illumination and color drift. We use a two-tier calibration design. At the 3D level, a Generative-Latent-Optimization (GLO)~\cite{bojanowski2018glo,martinbrualla2021nerfw} module learns a low-dimensional appearance code for each frame and feeds it to an MLP that predicts per-Gaussian color offsets, compensating for exposure changes, white-balance drift, and view-dependent appearance variations. The final MLP layer is initialized to zero, so early training starts from the original Gaussian colors. At the 2D level, a luminance-guided bilateral grid~\cite{chen2007bilateral} applies pixel-wise affine corrections to the rendered image after rasterization, handling residual tone mapping, vignetting, and spatially non-uniform brightness. The two modules are optimized jointly: GLO corrects appearance in 3D space, while the bilateral grid corrects image-space deviations with total-variation regularization.

    \item \textbf{MCMC-style densification.}
    Standard 3DGS densification based on screen-space gradient thresholds is unstable under sparse views, textureless regions, reflections, and occlusions. We adopt stochastic adaptive density control~\cite{kheradmand2024mcmc}, where low-contribution Gaussians are pruned by opacity, new Gaussians are sampled from surviving primitives, and opacity-modulated Langevin perturbations move inefficient Gaussians or let them fade out. This improves robustness on challenging surfaces such as polished floors and glass facades, while supporting adaptive allocation under a fixed Gaussian budget.

    \item \textbf{Transient suppression.}
    Dynamic objects violate the static-scene assumption required by 3DGS. We suppress them using 2D semantic or inpainting masks to down-weight dynamic regions, together with a learnable 3D mask scalar attached to each Gaussian. Multi-view inconsistent primitives are periodically identified and pruned, preventing dynamic-object contamination and providing cleaner static geometry for view synthesis and iterative repair.
\end{itemize}

Together, these modules enable the same \abotgs{} engine to handle generated panoramic videos and large-scale real-world captures with minimal scene-specific tuning. In experiments, the system reconstructs over $10^5$~m$^2$ of contiguous indoor and outdoor Insta360 X5 captures, robustly handling illumination drift, reflections, and dynamic clutter.

\paragraph{Reconstruction speedup}
We accelerate reconstruction in two ways. First, FLUX repair is applied only to a subset of training views selected by pose-based farthest-point sampling, reducing inference cost while preserving view coverage. Second, 3DGS training uses SH degree~0, a modified parallel FasterGS backend~\cite{hahlbohm2026faster}, and weight reuse across repair-and-refit passes, where each new round is initialized from the previous 3DGS. On a $4\times$4090 node, FLUX repair takes about 3 minutes per scene and each 3DGS round about 1 minute. The full pipeline, with three 3DGS rounds and two FLUX repair rounds, finishes in about 10 minutes including overhead. The resulting 3DGS assets can be further exported at multiple levels of detail via continuous LOD~\cite{clodgs2026}, enabling bandwidth-adaptive streaming and real-time rendering on heterogeneous client devices.
\subsection{Physical Attributes and Rendering Effects}
\label{sec:method_physics}

To convert the reconstructed 3DGS into a usable interactive world, \method{} extracts simple yet sufficient physical attributes and ships two signature rendering effects.

\paragraph{Occupancy and collision mesh}
Recent works such as SuGaR~\cite{guedon2024sugar}, 2DGS~\cite{huang20242dgs}, and PGSR~\cite{chen2024pgsr} extract high-quality meshes from 3D Gaussians by regularizing splats toward surfaces, but they assume clean multi-view captures with accurate camera poses and controlled lighting. Our setting differs: the input is a generative panoramic video whose Gaussians may contain floaters, density gaps, and illumination drift. We therefore opt for a volumetric occupancy approach that is robust to noisy Gaussian fields.
Specifically, we construct an automated pipeline from the Gaussian field to geometric collision assets.
We first compute an axis-aligned bounding box (AABB) from the camera trajectory and expand it by a preset margin to obtain a stable, well-shaped playable region.
Inside this AABB we perform Gaussian-aware voxelization of the 3DGS field: anisotropic densities are accumulated under a Beer--Lambert model, and after a morphological filter the occupied/empty boundary is extracted as a watertight surface mesh registered to the same coordinate frame as the Gaussians.

For collision geometry we do not rely on the occupancy mesh alone.
We explicitly build an ``air wall'' boundary shell---six dense, thin walls on the inner faces of the AABB---to prevent the camera from leaving the scene.
This shell is merged with the occupancy surface into a complete collision mesh.
To keep the collider lightweight and runtime-efficient, we retain only the largest connected components of the occupancy part, decimate the mesh, and apply Laplacian smoothing, yielding a compact navigation collider.
The initial camera pose is aligned with the initial panoramic viewpoint, with a safety margin reserved around it.
Camera traversal speed and collision radius are automatically derived from the AABB scale and adapt to scene size, ensuring consistent navigation feel across different scales.

\paragraph{Portal time-travel}
The portal effect underpins our flagship product surface. A portal is implemented as a per-pixel rendering mask that, within an oriented surface inside the scene, swaps the current Gaussian world for a companion one (\eg{} ``today's Red Cliffs'' $\rightarrow$ ``Battle of Red Cliffs''). Because both worlds share \method{}'s metric frame and the portal mask is computed in screen space, transitions remain visually seamless and computationally cheap.

\section{Evaluation}
\label{sec:eval}

We first showcase two capabilities that define \method{}'s design philosophy: input unification across all supported modalities (\cref{sec:eval_unified}) and a tripartite exploration-trajectory framework that governs how the virtual camera navigates 3D scenes (\cref{sec:eval_traj}). We then evaluate along four complementary dimensions: component-level validation of the panoramic video generator and our 3D Reinforcement Learning (3DRL) post-training stage (\cref{sec:eval_pano}) , component-level validation of the \abotgs{} reconstruction-and-repair engine (\cref{sec:eval_gs}), end-to-end generation quality against the strongest open- and closed-source baselines (\cref{sec:eval_e2e}), and system-level applicability on dimensions that matter for a consumer map product (\cref{sec:eval_system}). Following the protocol of HY-World 2.0~\cite{hyworld22026}, our primary baselines are Marble~\cite{marble2025worldlabs} and HY-World 2.0 itself.

\begin{figure}[t]
    \centering
    \includegraphics[width=1.0\textwidth]{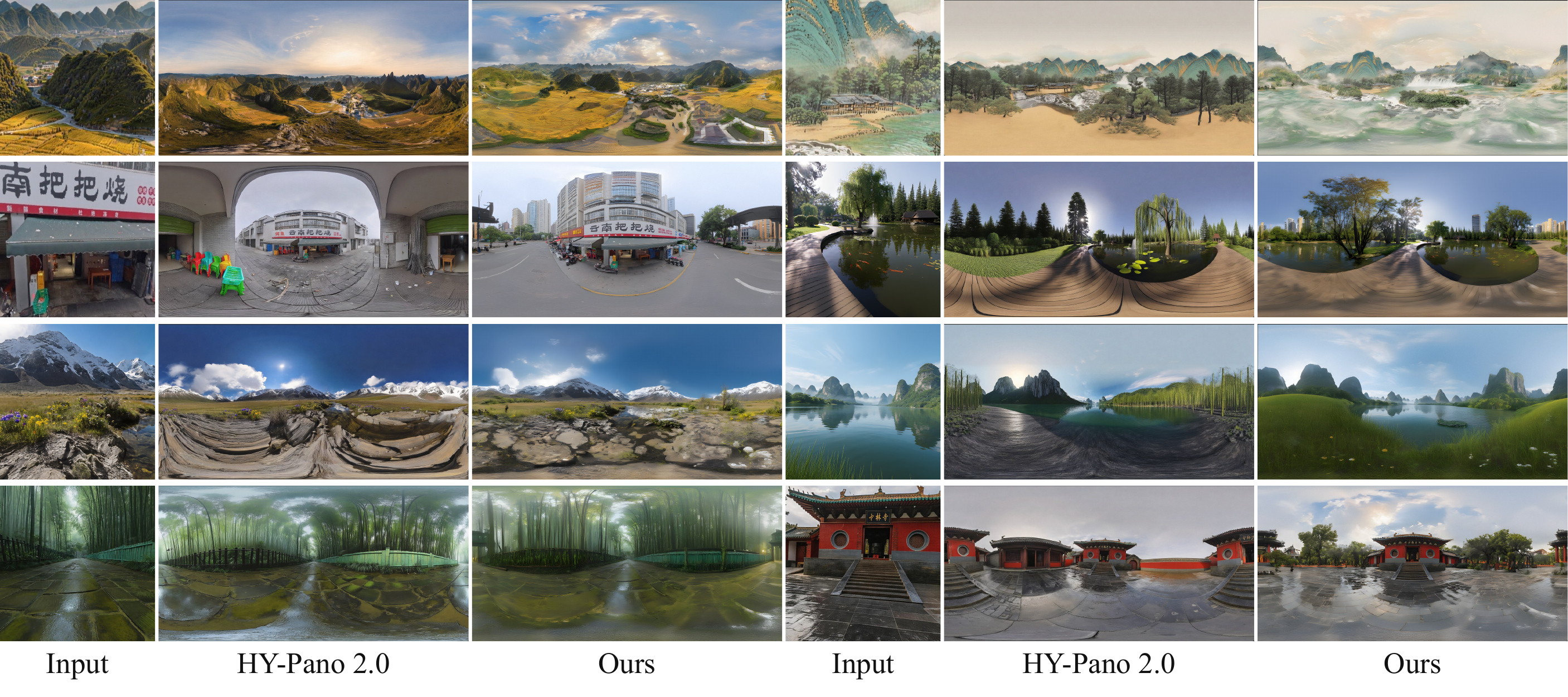}
    \caption{Single-image to panorama comparison. Each row shows, from left to right, the perspective input image, the equirectangular panorama produced by HY-Pano 2.0~\cite{hyworld22026}, and the panorama produced by \method{}, all generated from the same input. The two systems differ in how they extend the input's illumination, color tone, and texture statistics across the full $360^\circ$ surround.}
    \label{fig:panocompare}
\end{figure}

\subsection{Unified Framework}
\label{sec:eval_unified}

A defining property of \method{} is that every supported input modality---text prompts, single images, sparse multi-view photo sets, and casual video---is lifted into a common Spatial Generative Primitive (\cref{sec:method_input}) before any downstream generation occurs. This unification is what lets a single panoramic video generator and a single 3DGS reconstruction engine serve every input path, with no modality-specific branches. Below we show how this lifting is instantiated for two input modalities of increasing complexity: a single image and multi-view photographs.

\begin{figure}[!t]
\centering
\includegraphics[width=1.0\textwidth]{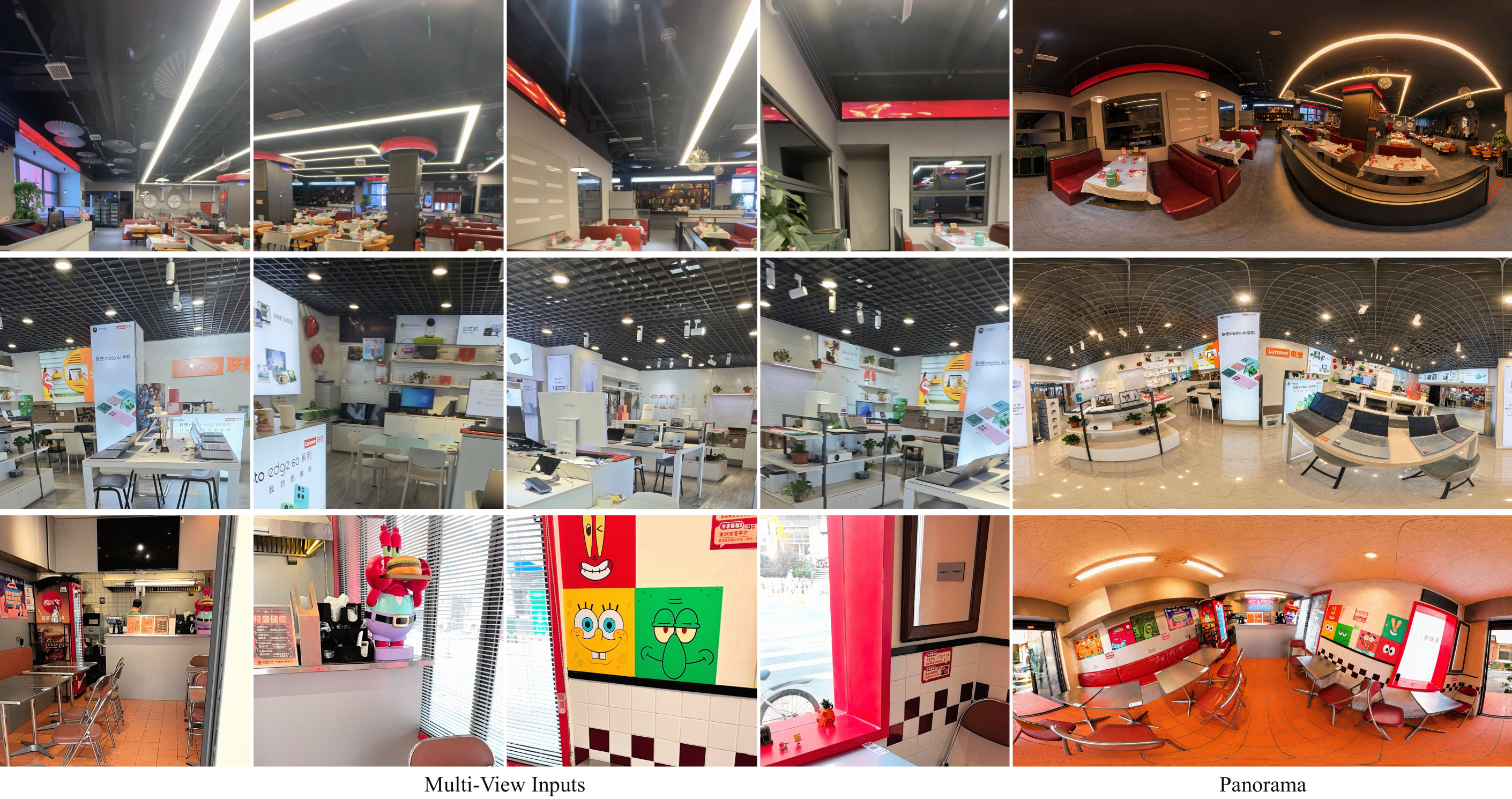}
\caption{Multi-view images to panorama. Each row shows four overlapping perspective input photos of an indoor scene (top: restaurant; middle: retail showroom; bottom: appliance store) and the equirectangular 360$^\circ$ panorama produced by the \method{} SGP lifting stage. The panorama recovers the full surround from sparse views without visible seams at the $\theta=\pm 180^\circ$ wrap-around, supplying the canonical appearance consumed by the downstream panoramic video generator.}
\label{fig:mv2pano}
\end{figure}

\paragraph{Single-image to panorama}
For the single-image input path, our panorama generator lifts a single perspective photo into a full $360^\circ$ equirectangular panorama that seeds the downstream Spatial Generative Primitive. \cref{fig:panocompare} places our panorama next to HY-Pano 2.0's~\cite{hyworld22026} on the same perspective inputs. ABot-Pano achieves higher conditional fidelity than HY-Pano 2.0. Its outputs more faithfully preserve the illumination, color tone, and artistic style of the input perspective image, with fewer instances of semantic drift. This improvement is likely attributable to the adaptive condition injection mechanism, which injects residual latents as strong spatial anchors and alleviates the information bottleneck and prior bias introduced by language-mediated processing. As a result, ABot-Pano provides more faithful style transfer and tighter semantic alignment. Meanwhile, it exhibits a slight decline in polar-region texture quality and open-scene expansion diversity, suggesting a trade-off between controllability and generative flexibility.

\paragraph{Multi-view images to panorama}
\cref{fig:mv2pano} shows three representative conversions. In each row, four overlapping perspective photos of an indoor scene---a restaurant, a retail showroom, and an appliance store---are lifted into a single equirectangular panorama covering the full $360^\circ$ surround. The resulting panoramas preserve the photometric detail of the inputs and exhibit no visible seam at the $\theta=\pm 180^\circ$ wrap-around.

Compared with direct monocular panorama generation or naive image stitching, our explicit SfM-plus-Gaussian reconstruction simultaneously delivers higher geometric precision, stronger cross-view consistency, more complete scene recovery and higher visual quality, thanks to a shared metric Gaussian field jointly optimized over all input observations.

\begin{figure}[!t]
\centering
\includegraphics[width=1.0\textwidth]{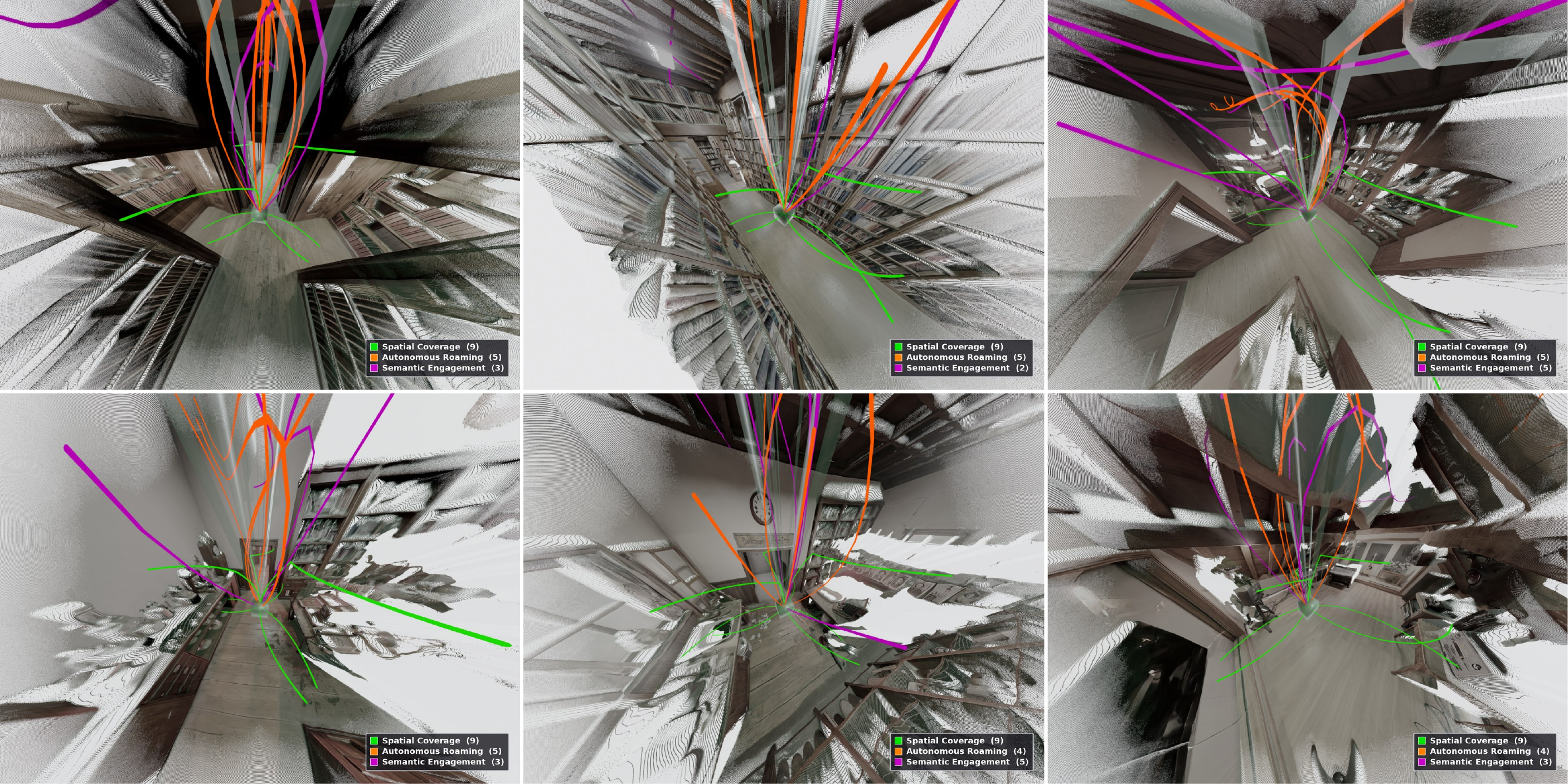}
\caption{Three exploration trajectory strategies. \emph{Spatial Coverage} (green, 9 paths) systematically traverses all navigable regions through geometrically grounded sampling, ensuring comprehensive observation of occluded architectural elements (corners, doorways, ceiling structures) while maintaining optimal parallax and view overlap. \emph{Autonomous Roaming} (orange, 5 paths) demonstrates adaptive navigation in unstructured environments, dynamically adjusting to geometric constraints (narrow corridors, cluttered furniture) without predefined semantic targets. \emph{Semantic Engagement} (purple, 5 paths) exhibits object-centric behavior, approaching salient entities (bookshelves, display cases) at information-rich distances and viewing angles determined by vision-language queries.}
\label{fig:traj_vis}
\end{figure}


\subsection{Exploration Trajectory}
\label{sec:eval_traj}

Beyond input unification, \method{} supports three complementary exploration strategies that govern how the virtual camera navigates a 3D scene to capture the multi-view evidence fed into downstream reconstruction. \cref{fig:traj_vis} illustrates these strategies.

\paragraph{Tripartite trajectory design}
Spatial Coverage trajectories ensure geometric completeness by systematically sampling navigable regions, eliminating blind spots that uniform path planners typically miss (corners, doorways, ceiling structures) while maintaining optimal parallax and view overlap for high-quality reconstruction. Autonomous Roaming trajectories handle environmental complexity through organic, context-aware motion patterns that adapt to geometric constraints (narrow corridors, cluttered furniture) without relying on semantic cues, establishing foundational spatial understanding before object engagement. Semantic Engagement trajectories maximize perceptual fidelity by locking onto target centroids at information-rich distances and viewing angles, generating high-fidelity multi-view observations that preserve fine-grained semantic details often lost in coverage-focused approaches.

This tripartite design resolves the inherent tension between exploration efficiency and reconstruction quality through context-aware adaptation. Unlike monolithic planners that prioritize geometric coverage or semantic targeting in isolation, \method{} dynamically balances all three objectives: spatial coverage ensures geometric completeness, autonomous roaming handles unstructured environments, and semantic engagement maximizes perceptual fidelity. Trajectories emerge from scene-specific geometric and semantic reasoning rather than fixed motion primitives, enabling robust performance across diverse architectural layouts---from confined hallways to open-plan spaces---while avoiding over-constrained or degenerate paths.

\begin{figure}[!t]
\centering
\includegraphics[width=0.9\textwidth]{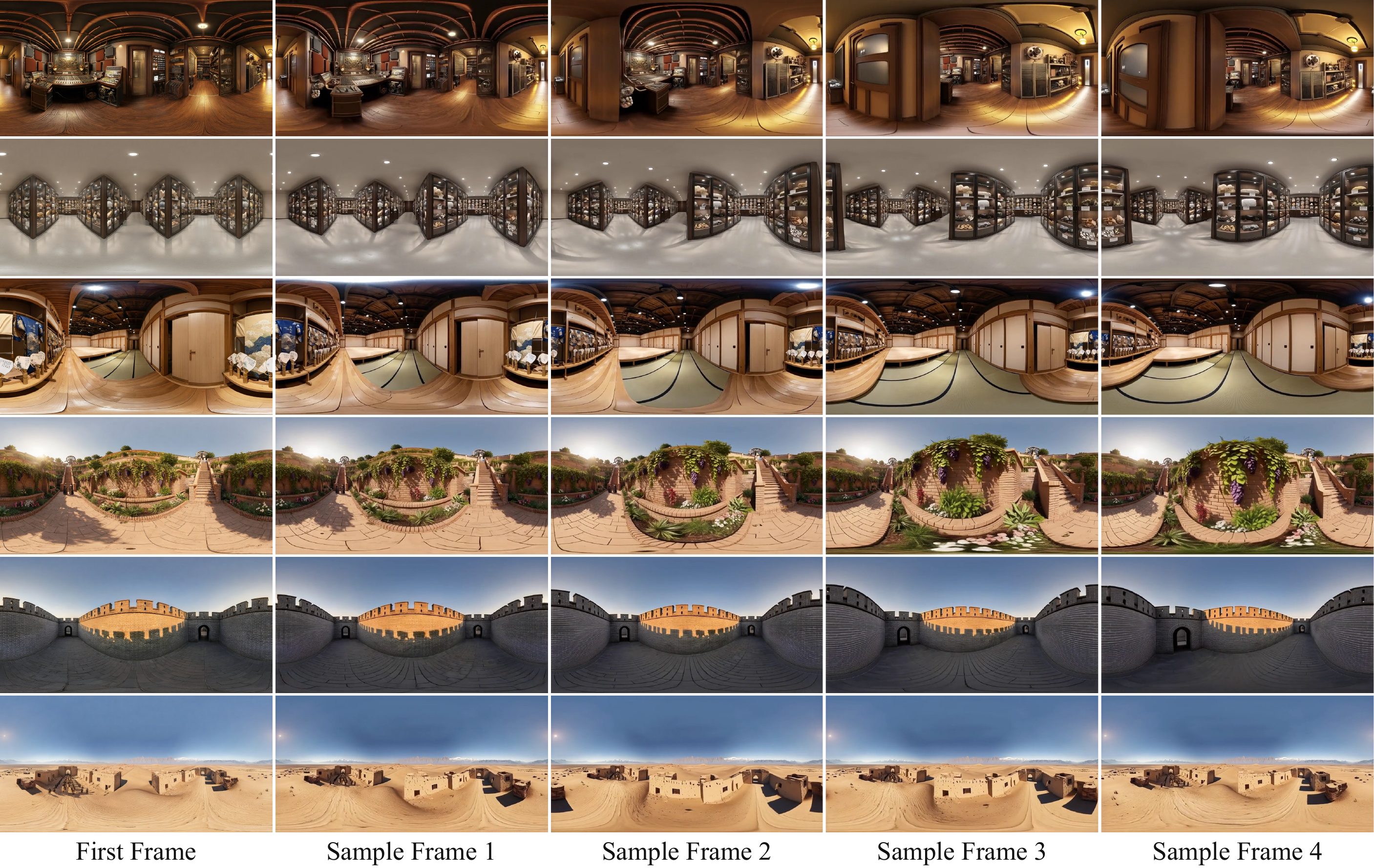}
\caption{Representative panoramic exploration videos generated by \method{} on indoor and outdoor scenes. Selected equirectangular frames along each virtual-camera trajectory exhibit coherent spatial structure, consistent frame-to-frame geometry, and fine-grained texture preservation across the full $360^\circ$ field of view.}
\label{fig:panovideogen}
\end{figure}

\begin{figure}[!t]
\centering
\includegraphics[width=0.9\textwidth]{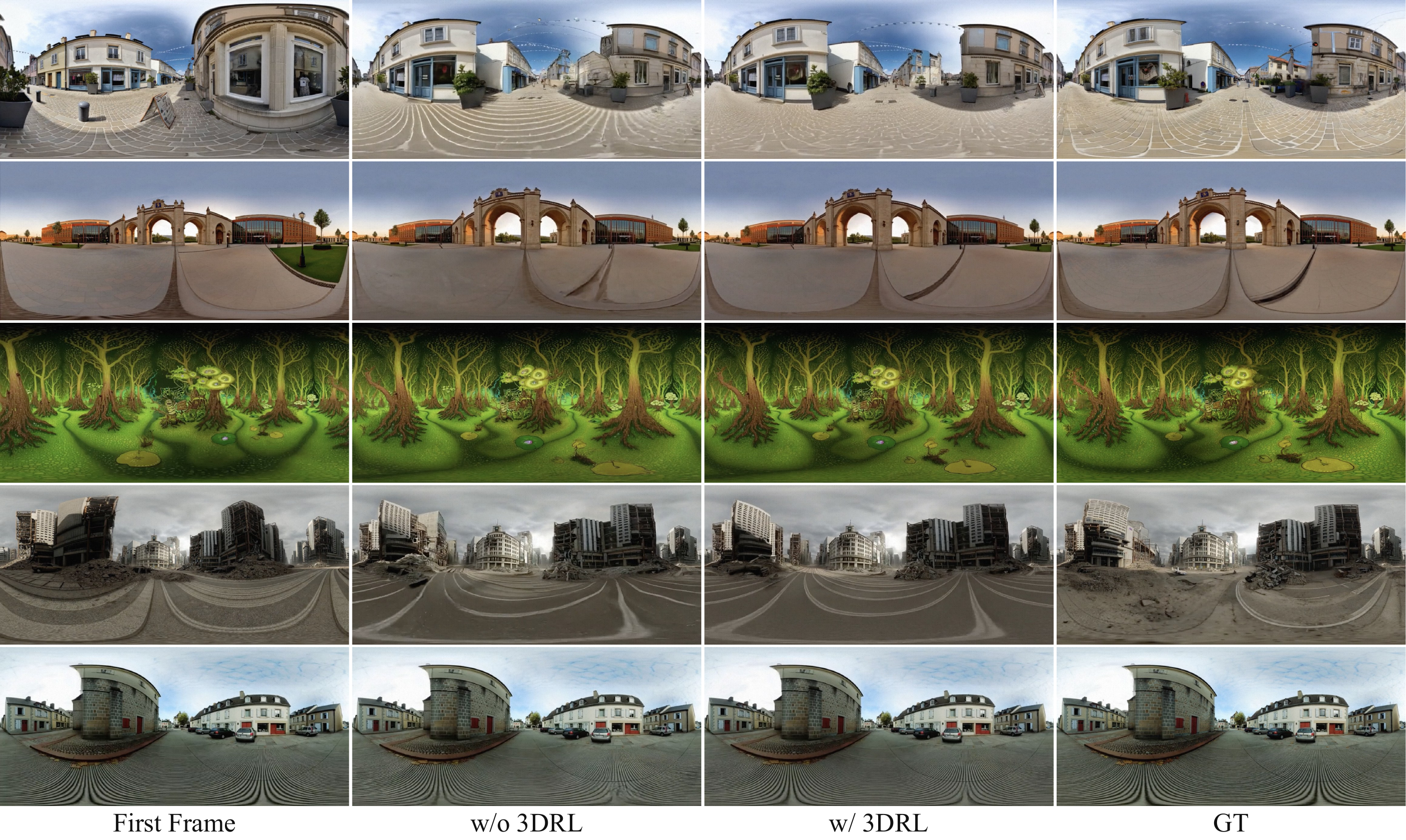}
\caption{Qualitative comparison of panoramic video generation on unseen test scenes. Each row shows, from left to right, the input first frame, the generation without 3DRL, the generation with 3DRL, and the ground truth. 
The proposed 3DRL stage improves geometric consistency and temporal coherence while preserving scene structure and visual fidelity.}
\label{fig:3drl_comparison}
\end{figure}

\begin{table}[!t]
\centering
\small
\caption{Quantitative comparison between without 3DRL and with 3DRL on 3D consistency and video quality.}
\label{tab:lora_eval}
\begin{tabular}{llcccc}
\toprule
Category & Metric & w/o 3DRL & w/ 3DRL \\
\midrule
\multirow{4}{*}{3D Consistency}
 & PSNR $\uparrow$                & 34.11 & \textbf{35.30}  \\
 & SSIM $\uparrow$               & 0.942  & \textbf{0.951}   \\
 & LPIPS $\downarrow$             & 0.089  & \textbf{0.082}   \\
 & Confidence $\uparrow$         & 0.396  & \textbf{0.412}  \\
\midrule
\multirow{5}{*}{Video Quality}
 & Aesthetic Quality $\uparrow$     & 46.74          & \textbf{47.85} \\
 & Imaging Quality $\uparrow$       & 43.34 &  \textbf{44.12}    \\
 & Motion Smoothness $\uparrow$     & 99.17          & \textbf{99.21} \\
 & Subject Consistency $\uparrow$   & 89.42          & \textbf{89.94} \\
 & Background Consistency $\uparrow$ & 94.60          & \textbf{94.61} \\
\bottomrule
\end{tabular}
\end{table}

\subsection{Panoramic Video Generation}
\label{sec:eval_pano}

The panoramic video generator is the perceptual core of \method{}: it produces the equirectangular video that the reconstruction engine later lifts into a 3DGS world. We isolate its two algorithmic levers---the base architecture (VACE 14B backbone with point-cloud rendering conditioning, latent circular padding, and LightX2V 4-step distillation; \cref{sec:method_video}) and the 3DRL post-training stage (\cref{sec:method_3drl})---and report how each contributes to the metrics that matter downstream.

\paragraph{Qualitative visualizations}
\cref{fig:panovideogen} presents representative panoramic exploration videos generated by \method{} on indoor and outdoor scenes. Across both settings, the generated equirectangular sequences exhibit coherent spatial structure and consistent frame-to-frame geometry while preserving fine-grained texture throughout the exploration trajectory. Compared with perspective-view video generation pipelines that typically produce and fuse multiple narrow-FOV clips, our panoramic formulation covers the full $360^\circ$ field of view in every frame, achieving equivalent spatial coverage with fewer frames. The dense cross-view overlap inherent in consecutive panoramic frames further suppresses view-inconsistency artifacts and provides richer multi-view supervision for the downstream \abotgs{} reconstruction.


\paragraph{3DRL post-training}
We select 100 held-out indoor and outdoor scenes to evaluate the generalization of 3D RL on unseen scenes. For each scene, we generate videos before and after RL using the same panorama initialization, condition video, and random seed, so that the comparison isolates the effect of RL training.

We evaluate the generated videos from two aspects: 3D consistency and video quality. For \emph{3D consistency}, we fit a 3DGS representation to each generated video. DA3~\cite{da3depth2025} is used to initialize the 3D Gaussians, which are then optimized for 1000 steps and rendered back to the original camera trajectory. We compare the rendered reconstruction with the generated video using PSNR, SSIM, LPIPS, and DA3 depth confidence. For \emph{video quality}, we report VBench~\cite{huang2024vbench} aesthetic quality, imaging quality, motion smoothness, subject consistency, and background consistency.

As shown in \cref{tab:lora_eval}, introducing 3DRL consistently improves both 3D consistency and video quality. For 3D consistency, PSNR increases from 34.11 to 35.30, SSIM improves from 0.942 to 0.951, LPIPS decreases from 0.089 to 0.082, and the confidence score also rises from 0.396 to 0.412. These results indicate that 3DRL enhances the geometric consistency and 3D reconstructability of generated videos. Meanwhile, all video quality metrics, including aesthetic quality, imaging quality, motion smoothness, subject consistency, and background consistency, are also improved. This suggests that 3DRL strengthens 3D consistency without degrading the overall visual quality or temporal stability of the generated videos.

\cref{fig:3drl_comparison} shows a representative side-by-side comparison on unseen test scenes. Compared with the w/o 3DRL result, the w/ 3DRL generations exhibit more stable scene structure---the geometric layout of buildings, roads, and foreground objects is more coherent across frames, with fewer local distortions and unnatural deformations. The 3DRL outputs are visibly closer to the ground-truth panoramas, confirming that the 3D-aware reward encourages geometrically and temporally consistent panoramic video generation without sacrificing visual quality.

\begin{figure}[!t]
\centering
\includegraphics[width=1.0\textwidth]{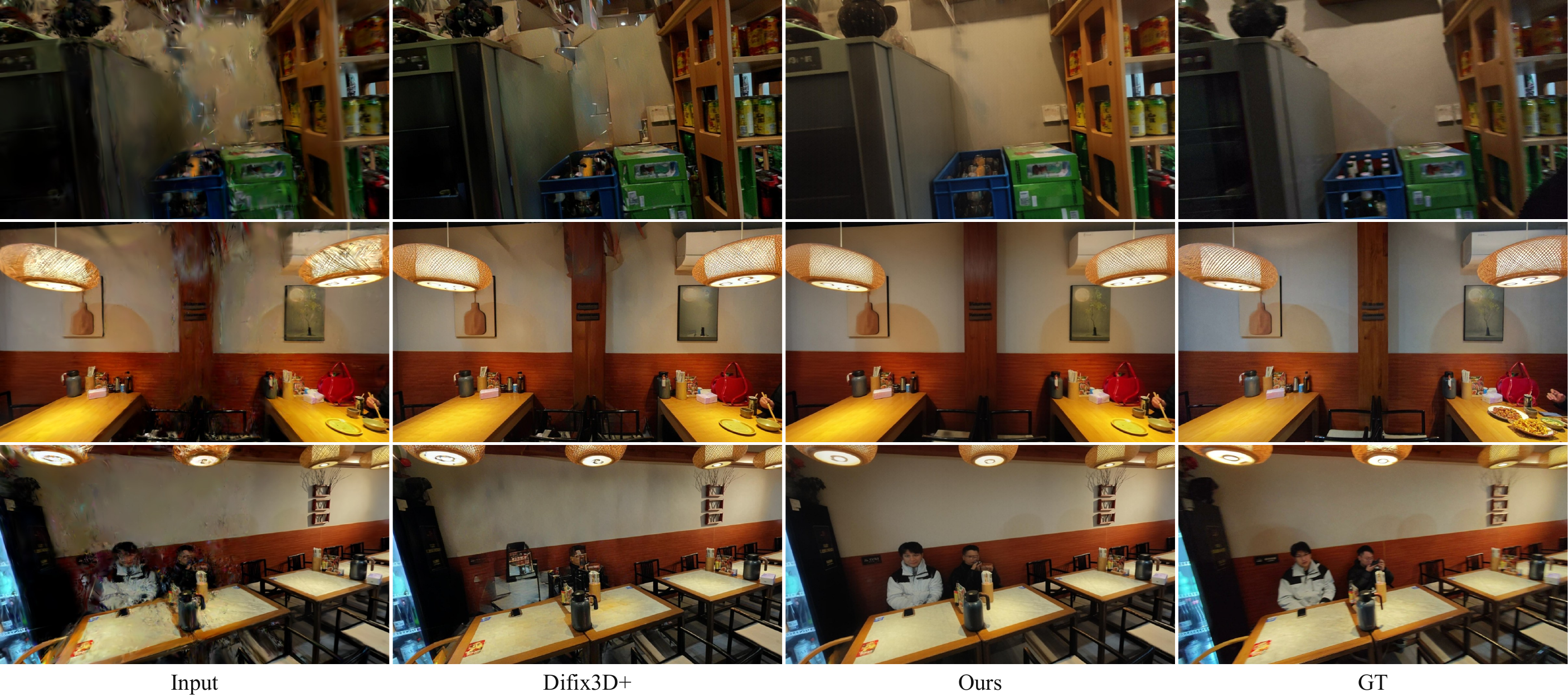}
\caption{Qualitative comparison of 3DGS repair on side-view test renders held out from sparse-view forward reconstruction. Each scene shows four columns left-to-right: \emph{Input}, the defective 3DGS render at the held-out side view (wall distortion, texture stretching, local holes, and broken person regions); \emph{Difix3D+}~\cite{difix3dplus2025} under the same reference-view conditioning (removes some artifacts but over-smooths complex regions such as persons and tabletops); \emph{Ours}, our FLUX-based repair under the same input and reference view (suppresses artifacts while preserving scene geometry and semantic content); and \emph{GT}, the real image captured at the side test view, used as the reference for both quantitative evaluation and visual comparison.}
\label{fig:difix_vis}
\end{figure}

\begin{figure}[!t]
\centering
\includegraphics[width=0.92\textwidth]{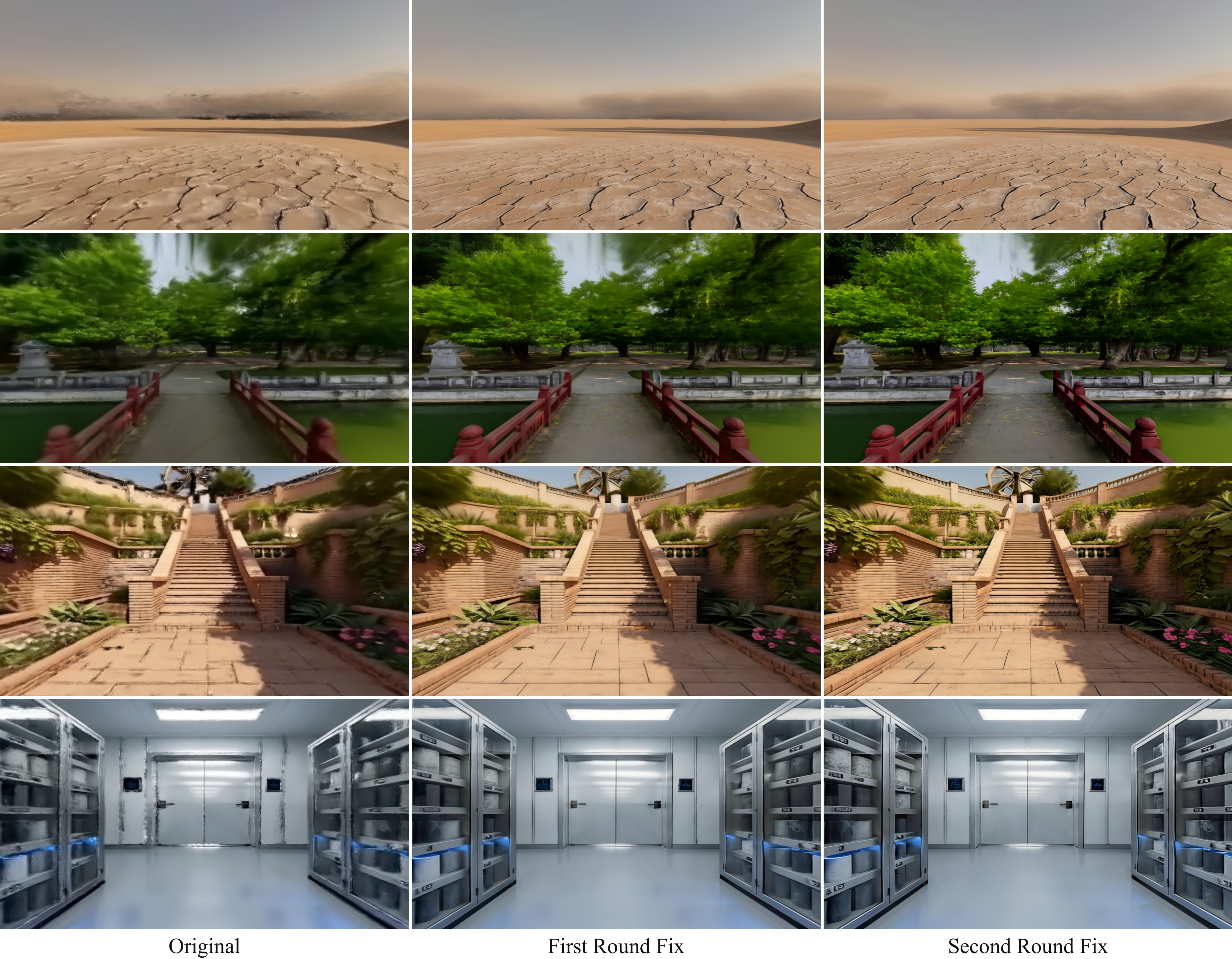}
\caption{Qualitative comparison of iterative FLUX-based Gaussian repair. The first round fix significantly improves the original reconstruction by removing severe photometric and geometric inconsistencies and enhancing resolution. The second round fix further suppresses the remaining geometric and photometric drift, resulting in sharper textures and more stable details.}
\label{fig:multi_round_3dgs}
\end{figure}

\subsection{3DGS Optimization}
\label{sec:eval_gs}

The second component-level study validates the \abotgs{} reconstruction-and-repair engine (\cref{sec:method_gs}). To isolate the repair component from the rest of the pipeline, we first evaluate it on a controlled real-world benchmark where 3DGS representations are deliberately constructed with insufficient view coverage, comparing directly against the Difix3D+ baseline~\cite{difix3dplus2025}. Then we present gaussian results by the iterative FLUX-based repair. Finally, we show the robustness of our results on reflective surfaces.

\begin{figure}[!t]
\centering
\includegraphics[width=0.92\textwidth]{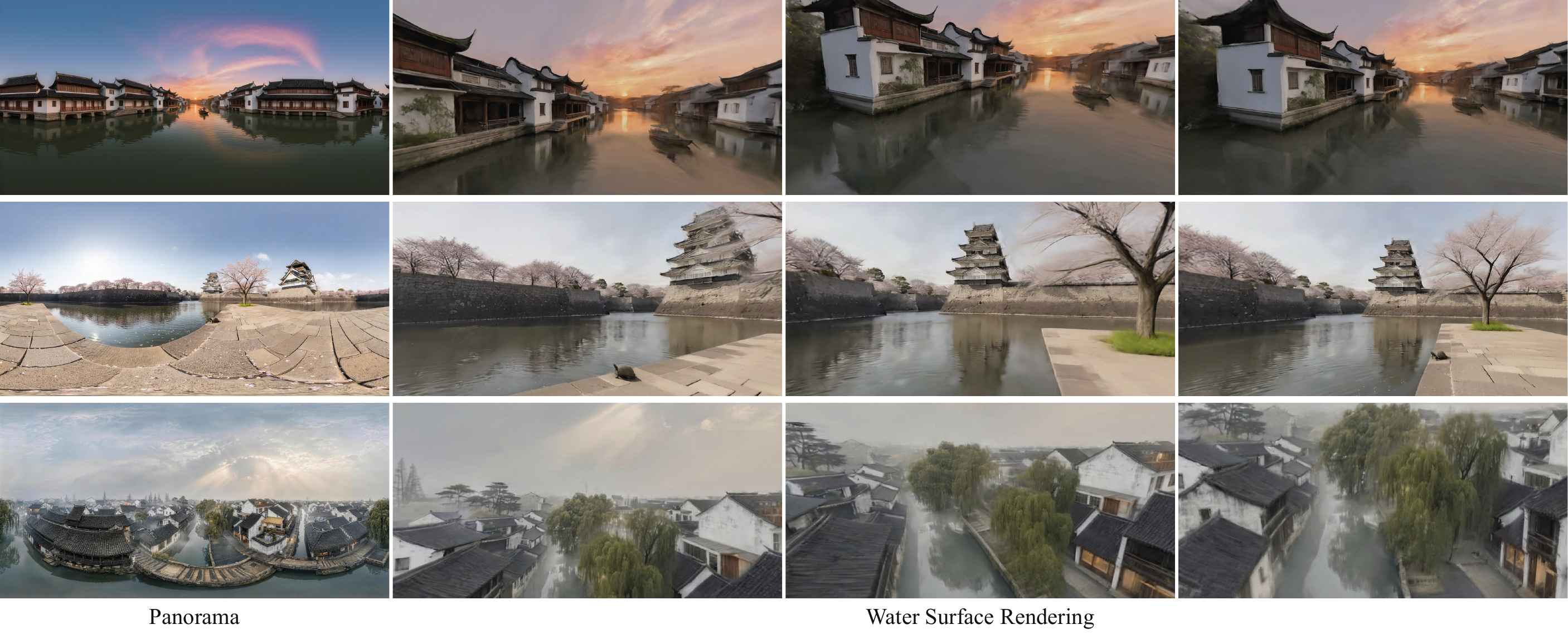}
\caption{Robustness to reflective surfaces. Left: the input panorama of a traditional architectural scene dominated by water bodies. Right: 3DGS renders from multiple viewpoints of three water-heavy scenes. Water-surface reflections remain stable across viewpoints, with no over-proliferation of Gaussians or illumination distortion.}
\label{fig:3dgs_robustness}
\end{figure}

\begin{table}[t]
\centering
\small
\caption{Quantitative comparison with Difix3D+ on 3D consistency metrics. Bold marks the better value per column.}
\label{tab:difix_vs_ours}
\begin{tabular}{lccc}
\toprule
Method & PSNR $\uparrow$ & SSIM $\uparrow$ & LPIPS $\downarrow$ \\
\midrule
Difix3D+ & 20.2968          & 0.7184          & 0.2667          \\
Ours  & \textbf{21.6118} & \textbf{0.7926} & \textbf{0.2314} \\
\bottomrule
\end{tabular}
\end{table}

\paragraph{FLUX-based repair}
We collect real-world scenes with deliberately challenging test views: side-view perspectives that differ significantly from the forward-facing capture trajectory used during 3DGS reconstruction. For each scene, we reconstruct a 3DGS representation from only sparse forward-facing frames sampled from the capture, which leaves side regions under-constrained and produces visibly defective renders (wall distortion, texture stretching, local holes, and broken person regions). Repair is then performed on these defective side-view renders, with the same reference view supplied as conditional input across all methods for fairness. The repaired images are compared against side-view ground truth on PSNR, SSIM, and LPIPS. This protocol is substantially more demanding than a random train/test view split: sparse forward views cannot sufficiently constrain side-region geometry and appearance, and the large view gap requires the repair method to exploit reference information effectively and generalise across wide view changes.

\cref{tab:difix_vs_ours} reports the result. Our method outperforms Difix3D+ on all three metrics: PSNR rises from 20.30 to \textbf{21.61} (+1.31 dB), SSIM from 0.718 to \textbf{0.793} (+0.074), and LPIPS drops from 0.267 to \textbf{0.231} (-0.035). The magnitude of the SSIM and LPIPS gains---which weigh structural and perceptual similarity more heavily than pixel-wise PSNR---indicates that our method is doing meaningful structural repair rather than mere color re-balancing. Qualitatively (\cref{fig:difix_vis}), Difix3D+ removes some artifacts but tends to over-smooth complex regions, weakening or erasing person and tabletop details; our method suppresses artifacts while preserving scene geometry and semantic content, staying closer to ground truth and demonstrating stronger cross-view consistency under large view differences. \cref{fig:multi_round_3dgs} presents the original 3DGS results alongside those trained after two rounds of FLUX-based repair. Compared to the pre-repair baseline, the refined outputs exhibit notably improved sharpness and richer fine-grained details, as evidenced by the enhanced rendering of vegetation, ground textures, and specular light reflections.

\paragraph{Robustness to reflective surfaces}
\cref{fig:3dgs_robustness} stress-tests the 3DGS engine on a known failure mode of standard 3DGS: sparsely-textured reflective surfaces such as polished floors, glass facades, and, in this test, the water bodies that dominate the input panorama. Under screen-space-gradient densification, these surfaces corrupt the gradient signals that trigger Gaussian splitting and cloning, producing either local over-proliferation (visible as floating clouds above the surface) or under-coverage that distorts illumination and reflections. Our MCMC-style densification (\cref{sec:method_gs}) replaces the deterministic schedule with opacity-weighted probabilistic sampling, decoupling density control from gradient triggers; the right panels of \cref{fig:3dgs_robustness} show corresponding novel-view renders, where water-surface reflections remain stable and consistent across viewpoints without the artifacts described above.

\subsection{End-to-End Generation Quality}
\label{sec:eval_e2e}

We follow the end-to-end evaluation protocol of HY-World 2.0 Section 8.1.5~\cite{hyworld22026}. For each test prompt (a single perspective image or a short text description) we produce an equirectangular panorama and generate 3D worlds by each method with the same input for fairness and convenience of comparison. We then sample a fixed set of novel views, and report image-level perceptual metrics (Q-Align, CLIP-IQA+, Laion-Aes, CLIP-I) on the rendered output. We run all systems through their public APIs / open-source releases under default settings.

\begin{figure}[t]
\centering
\includegraphics[width=1.0\textwidth]{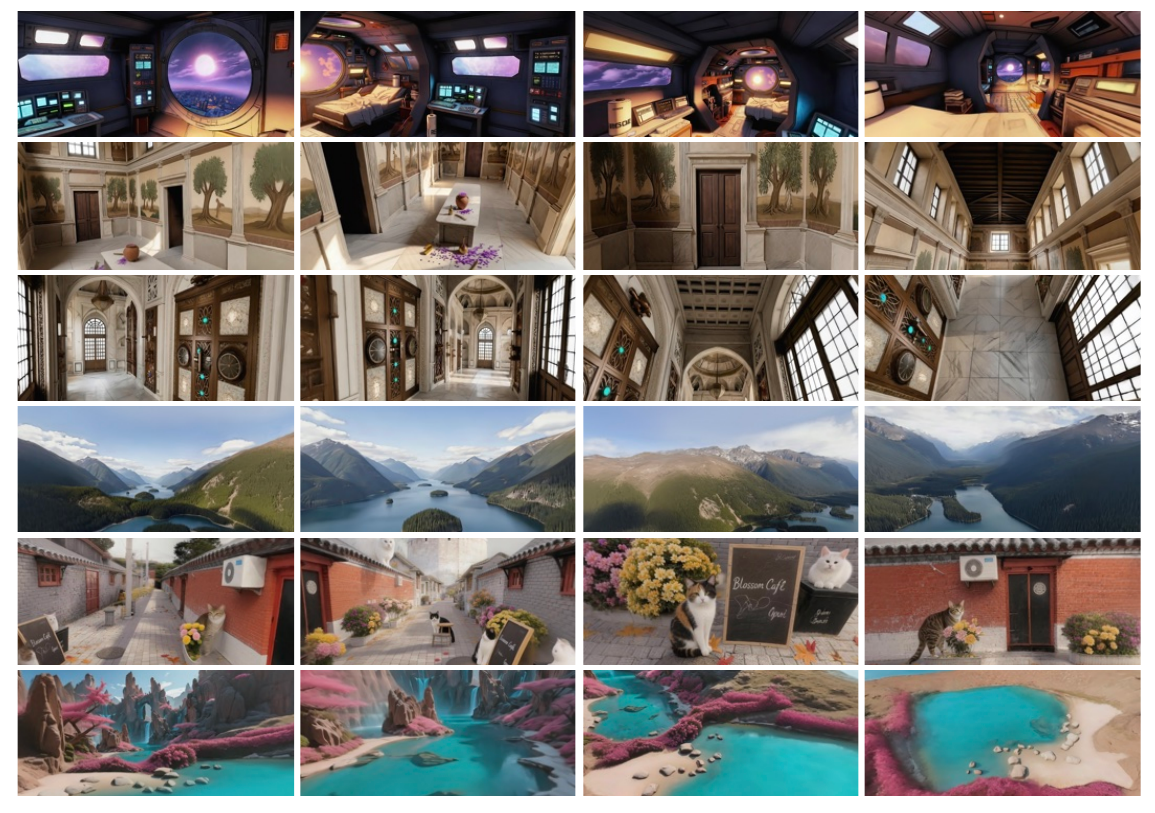}
\caption{Representative novel-view renders of 3DGS worlds produced by \method{} end-to-end across indoor, street-level outdoor, and aerial scene categories. The full pipeline---SGP extraction, 14B panoramic video generation with latent circular padding and 3DRL post-training, followed by iterative FLUX-based repair---yields stable geometry and sharp texture across all three regimes.}
\label{fig:ours}
\end{figure}

\paragraph{Qualitative visualizations}
\cref{fig:ours} presents representative novel-view renders of 3DGS worlds produced by \method{} end-to-end. Despite covering three visually distinct regimes---indoor, street-level outdoor, and aerial---all scenes are produced by the same unified inference path, with no scene-specific routing or specialized branches.

\begin{figure}[!t]
\centering
\includegraphics[width=1.0\textwidth]{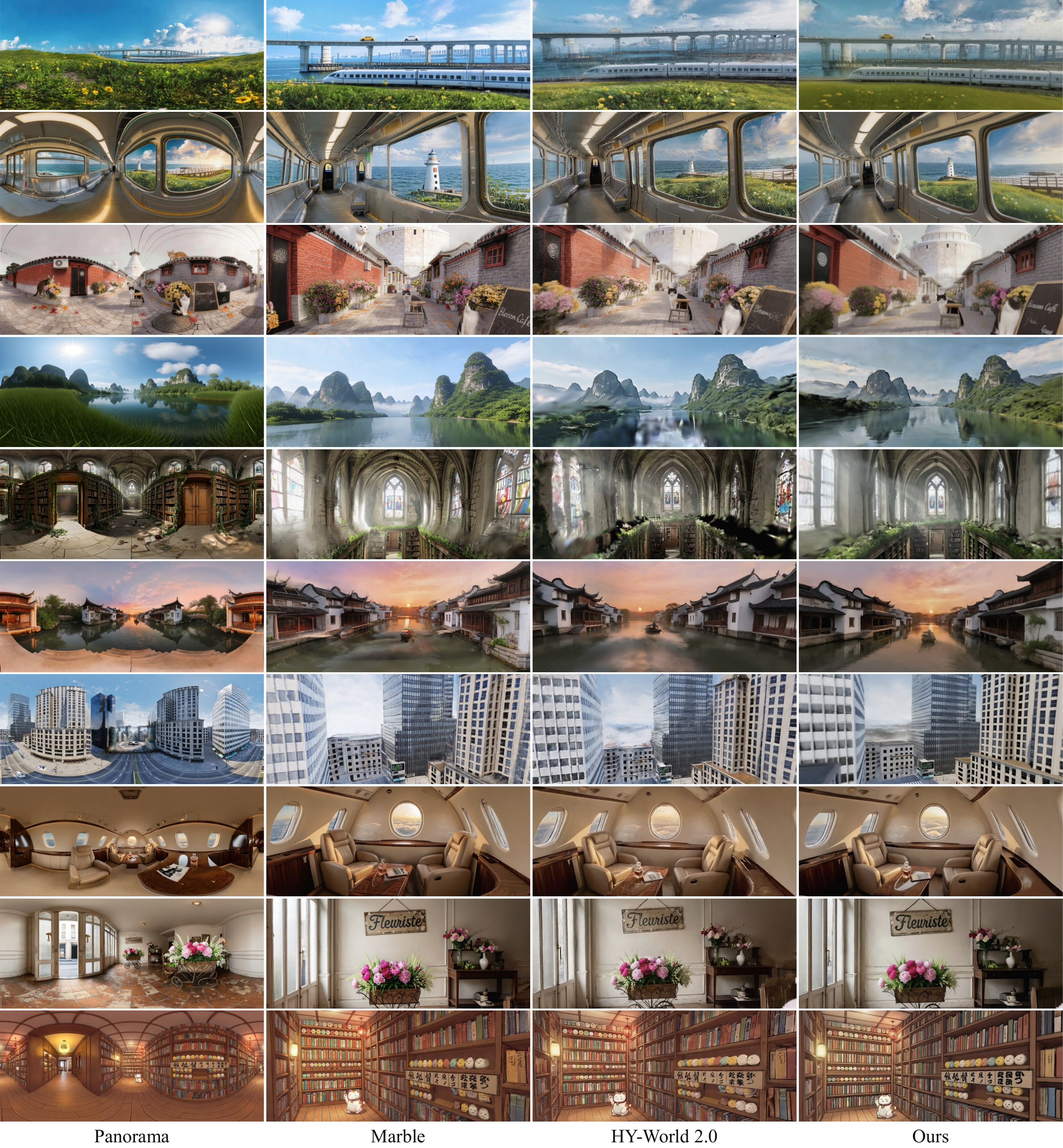}
\caption{Qualitative comparison of \method{} against Marble and HY-World 2.0 on indoor, street, scenic, and aerial scene categories.}
\label{fig:vs_marble}
\end{figure}

\paragraph{Quantitative comparison with HY-World 2.0}
\cref{tab:hy_vs_ours} reports four visual quality metrics—Q-Align, CLIP-IQA+, Laion-Aes, and CLIP-I—across Indoor, Outdoor, and All splits. The first three measure perceptual image quality while CLIP-I measures content fidelity to the input prompt. Our method achieves the strongest and most consistent gains on the two metrics that are most diagnostic for world generation. On Laion-Aes, our method improves over HY-World 2.0 by +0.1772 on Indoor, +0.1783 on Outdoor, and +0.1766 on the full split, indicating a uniform, distribution-wide aesthetic gain rather than a category-specific artifact, thanks to the 14B VACE backbone that—relative to the 1.3B backbone typical of panoramic baselines—retains substantially richer texture detail at the panorama's high effective resolution. On CLIP-I, our method likewise leads on every split, by +0.0062 on Indoor, +0.0093 on Outdoor, and +0.0069 on All; the Outdoor gain is the largest, consistent with the AMAP aerial-asset training distribution that biases our generator toward street, scenic, and aerial content. On the remaining two metrics the two systems are within close range and trade splits. HY-World 2.0 is marginally ahead on Q-Align. CLIP-IQA+ splits between the two systems: our method leads on Indoor and on All, while HY-World 2.0 leads on Outdoor. Taken together, the two systems are at parity on low-level perceptual quality, while our method exhibits clear and uniform advantages on aesthetic appeal and content fidelity to the input.

\begin{table}[t]
\centering
\small
\caption{Quantitative comparison with HY-World 2.0 on visual quality metrics across indoor, outdoor, and full splits. Bold marks the better value per cell.}
\label{tab:hy_vs_ours}
\begin{tabular}{llcccc}
\toprule
Val Set & Method & Q-Align $\uparrow$ & CLIP-IQA+ $\uparrow$ & Laion-Aes $\uparrow$ & CLIP-I $\uparrow$ \\
\midrule
\multirow{2}{*}{Indoor}
 & HY-World 2.0 & \textbf{4.1578} & 0.5412          & 5.3277          & 0.9129 \\
 & Ours         & 4.1503          & \textbf{0.5573} & \textbf{5.5049} & \textbf{0.9191} \\
\midrule
\multirow{2}{*}{Outdoor}
 & HY-World 2.0 & \textbf{3.7592} & \textbf{0.5220} & 5.1887          & 0.8970 \\
 & Ours         & 3.6962          & 0.5198          & \textbf{5.3670} & \textbf{0.9063} \\
\midrule
\multirow{2}{*}{All}
 & HY-World 2.0 & \textbf{4.0755} & 0.5348          & 5.3137          & 0.9105 \\
 & Ours         & 4.0426          & \textbf{0.5469} & \textbf{5.4903} & \textbf{0.9174} \\
\bottomrule
\end{tabular}
\end{table}

\paragraph{Qualitative comparison with Marble and HY-World 2.0}
\cref{fig:vs_marble} contrasts \method against Marble and HY-World 2.0 across indoor, street, scenic, and aerial categories. Overall, while Marble and HY-World 2.0 deliver competitive 3DGS reconstructions with high per-frame rendering quality, they both exhibit recurring failure modes—reconstruction blind spots in regions weakly covered by the conditioning inputs, localized blur on fine structures, and geometric distortion under viewpoints that deviate from the canonical capture pose. \method, in contrast, shows consistent advantages in coverage, blur resistance, and structural fidelity. As shown in \cref{fig:vs_marble}, our method produces noticeably sharper and more faithful results on challenging regions, including reflective water surfaces, on-scene text and signage, viewpoints with large parallax relative to the panorama origin, and texture-rich areas where competing pipelines lose high-frequency detail.

\begin{figure}[!t]
\definecolor{camgray}{RGB}{215,215,215}
\definecolor{camblue}{RGB}{83,117,192}
\definecolor{camorange}{RGB}{221,132,71}
\definecolor{campink}{RGB}{236,117,179}
\centering
\includegraphics[width=0.85\textwidth]{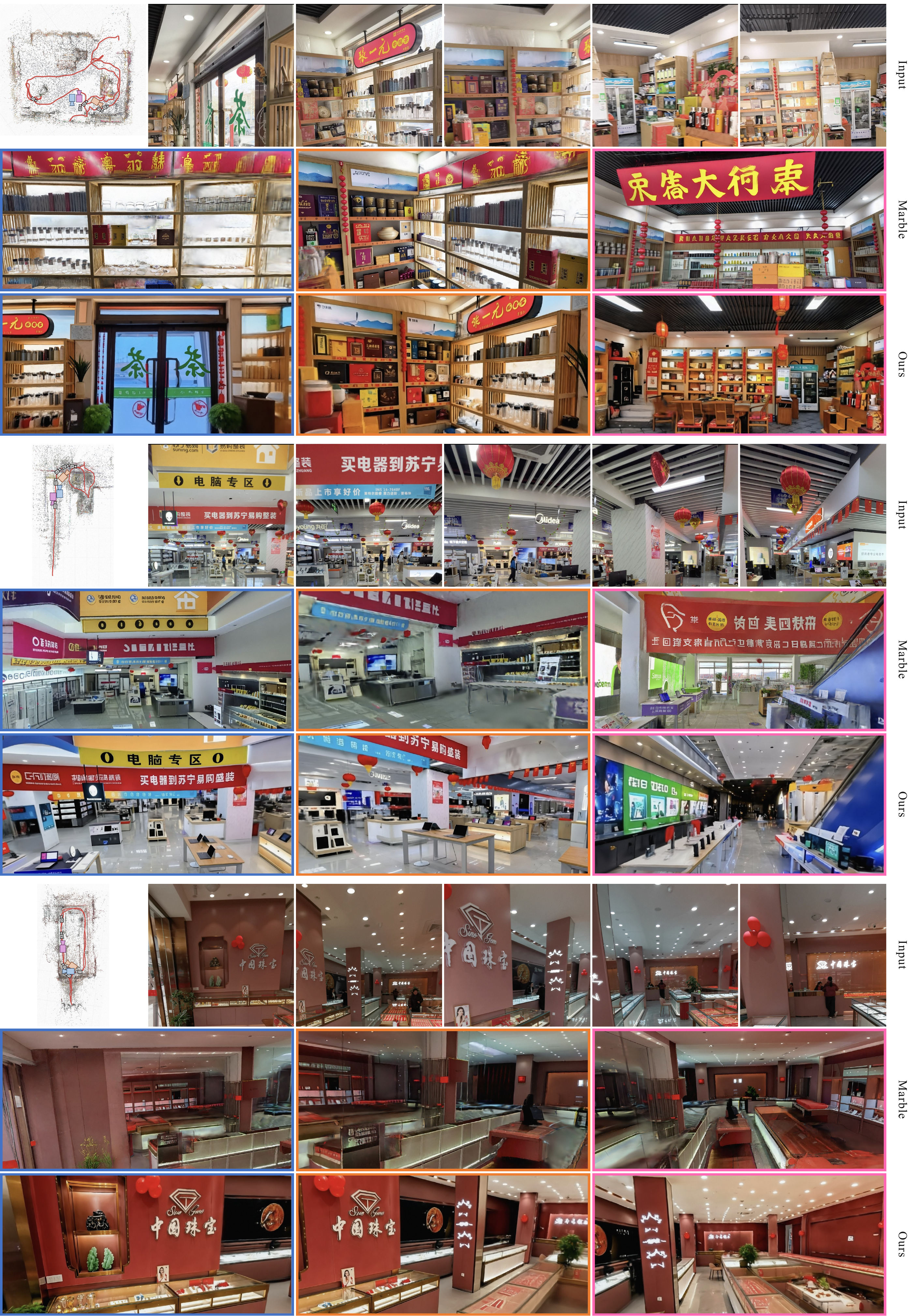}
\caption{Multi-view consistency under rich multimodal input. The first image in the Input row shows a top-down structural overview of the scene along with the approximate positions of input and rendered cameras, where numbered \textcolor{camgray}{gray} cameras denote input views and \textcolor{camblue}{blue}, \textcolor{camorange}{orange}, and \textcolor{campink}{pink} cameras indicate rendered viewpoints. The subsequent five images in the Input row correspond to the \textcolor{camgray}{gray} cameras respectively; The bottom two rows present rendering results from Marble and Ours, with \textcolor{camblue}{blue}, \textcolor{camorange}{orange}, and \textcolor{campink}{pink} borders matching their corresponding camera positions. \method{} achieves stronger fidelity to the input observations through unified SGP extraction compared with Marble.}
\label{fig:vs_marble_mv}
\end{figure}

\paragraph{Behavior under rich multimodal input}
The most pronounced gap between \method{} and Marble appears under rich multimodal inputs as shown in \cref{fig:vs_marble_mv}. When the input is a sparse multi-view photo set or a casual video, Marble seems to discard the original geometric and photometric evidence and visibly amplify hallucination. \method{} instead lifts such inputs into the SGP via a rigorous geometry pipeline that preserves metric and photometric evidence, yielding markedly lower hallucination and tighter alignment with the observed scene.

\subsection{System-Level Applicability}
\label{sec:eval_system}

Beyond pixel-level metrics, deploying a 3D world model in the consumer map ecosystem requires evaluating system-level properties.


\paragraph{Coverage}
\method{} is one of the few generative world models that consistently support the full indoor / ground-level outdoor / aerial spectrum within a single product surface. Marble is closed-source and dominates indoor scenes but offers limited city-scale coverage; HY-World 2.0 is open but optimized for game-asset and embodied-AI use cases.

\paragraph{Map and POI grounding}
Through the integration with our internal map service, every generated world is anchored to a geographic POI when the input is map-relevant (\eg{} a restaurant photo, a landmark image). This positions \method{} naturally as the spatial extension of a consumer map application, a property neither Marble nor HY-World 2.0 exposes.

\paragraph{Portal time-travel}
The portal time-travel product surface is unique to \method{}: neither Marble nor HY-World 2.0 expose a comparable consumer-facing experience. We argue that this is not an aesthetic choice but a structural one, enabled by the metric-aligned SGP and the cheap portal compositing mode of \cref{sec:method_physics}.

\clearpage

\clearpage

\paragraph{FPV landmark orbit}
For any landmark POI, \method{} generates a full $360^\circ$ scene that powers a first-person FPV orbiting experience. In our internal trials, FPV orbit qualitatively extends user viewing time relative to a single static panorama, consistent with the immersive-consumption pattern that short-video platforms have established.

\section{Conclusion}
\label{sec:conclusion}

We presented \method{}, a universal multimodal 3D world model that turns a single image, a text prompt, a sparse multi-view set, or a casual video into a navigable, photorealistic 3DGS world---spanning indoor, ground-level outdoor, and aerial regimes---within one unified pipeline. The design centers on the Spatial Generative Primitive as a compact canonical intermediate, panoramic video as an efficient exploration medium enhanced by 3D Reinforcement Learning, iterative FLUX-based repair for sharp and view-consistent reconstruction, and a purpose-built data engine that renders geometry-aligned training supervision from reconstructed 3DGS assets. Together these choices let a single system serve both faithful reality reconstruction from rich inputs and low-barrier creative generation from a single photograph or sentence. Experiments show that \method{} surpasses HY-World 2.0 on aesthetic quality and content fidelity, and demonstrates markedly stronger scene fidelity than Marble under rich multimodal inputs. Every generated world is further anchored to a real-world geographic point of interest, positioning \method{} as a map-native spatial exploration engine at consumer scale.

Looking forward, three directions are most promising: scaling SGP composition to city-scale environments while preserving cross-primitive geometric consistency; deeper integration with planet-scale 3D map services for continuous globe-to-room exploration; and enriching the physical layer---extending the current collision mesh and portal system toward richer interactive affordances such as relighting, dynamic elements, and agent interactions---to bring \method{} closer to a true ``explore any 3D space'' experience.

\clearpage
\section{Contributions}
\label{sec:contributions}

\textbf{Algorithm:} Mingchao Sun, Luyang Tang, Yu Liu, Xu Yan, Zhan Li, Yunwei Zhang, Fei Yu, Zengye Ge, Yumin Liu, Jiacheng Zhang, Yongchang Zhang, Jiawei Zhang, Zhicheng Liu, Zhongxu Sun, Tianjian Ouyang, Wenzheng Chen

\textbf{Engineering / Deployment:} Shixing Yang, Nianfei Fan, Guodong Sun, Huan Li, Zheng Zhou, Yongze Li, Yingliang Peng, Mengmeng Du, Yuan Liu

\textbf{Art \& Product Design:} Haozhe Shi, Chunnuo Gong, Chengzhen Yu, Chunxue Jia, Yang Liu, Shiying Zeng

\textbf{Project Sponsor:} Mu Xu, Junnan Lai, Baoquan Chen, Ning Guo

\textbf{Project Lead:} Hang Zhang, Hongyu Pan, Mingchao Sun

We would like to express our sincere gratitude to Yu Lei, Chong Sun, and Qianwei Wang for their valuable support and contributions to this project.

\bibliographystyle{IEEEtran}
\bibliography{reference}

\end{document}